\documentclass{article}
\usepackage{iclr2027_conference,times}
\usepackage[T1]{fontenc}
\usepackage{amsmath,amssymb,amsthm,mathtools}
\usepackage{booktabs,array,graphicx,tabularx,multirow,longtable}
\usepackage{algorithm,algpseudocode}
\usepackage{microtype}
\usepackage{placeins}
\usepackage{subcaption}
\usepackage{url}
\graphicspath{{images/}}
\usepackage{amsmath,amsfonts,bm}

\def\eqref#1{equation~\ref{#1}}

\def\1{\bm{1}}

\DeclareMathAlphabet{\mathsfit}{\encodingdefault}{\sfdefault}{m}{sl}
\SetMathAlphabet{\mathsfit}{bold}{\encodingdefault}{\sfdefault}{bx}{n}

\DeclareMathOperator*{\argmin}{arg\,min}

\usepackage{hyperref}
\hypersetup{
  pdftitle={Remember, Revisit, Revise: Scaffold Memory for Long-Horizon 3D Geometry},
  pdfauthor={Soumick Sarker and Mengmi Zhang},
  pdfsubject={Scaffold Memory for Revision (SMoRe)},
  bookmarksnumbered=true
}

\iclrfinalcopy

\newcommand{\smr}{\textsc{SMoRe}}

\title{Remember, Revisit, Revise: Scaffold Memory for Long-Horizon 3D Geometry}

\author{Soumick Sarker$^1$ \qquad Mengmi Zhang$^{1\dagger}$\\
\normalfont $^1$Deep NeuroCognition Lab, Nanyang Technological University, Singapore\\
\normalfont $^\dagger$Address correspondence to \texttt{mengmi.zhang@ntu.edu.sg}}

\begin{document}
\maketitle

\fancyhead{}
\renewcommand{\headrulewidth}{0pt}

\begin{abstract}

In 3D computer vision, geometry foundation models estimate camera poses and 3D scene structure from video sequences, but maintaining consistent reconstructions over long horizons remains challenging. Feed-forward models process long sequences as independent, fixed-length frame windows and subsequently align their predictions, causing pose and reconstruction errors to accumulate across windows. Streaming models retain information from previous frames to inform future predictions, but cannot revise earlier estimates when subsequent observations reveal errors. To address these limitations, we introduce \underline{S}caffold \underline{M}em\underline{O}ry for \underline{Re}vision (\smr{}), a framework that equips existing geometry foundation models with a revisable external memory for long-horizon inference. Inspired by hippocampal associative memory in biological brains, which links spatial context with sensory experience, \smr{} associates \emph{where} an observer is with \emph{what} it observes, enabling past observations to be stored, retrieved, verified against new evidence, and revised when inconsistencies are detected. Crucially, \smr{} integrates with existing geometry foundation models while keeping them frozen, requiring no additional training or fine-tuning. We evaluate \smr{} with eight competitive geometry models on five datasets spanning camera pose estimation, multi-view depth estimation, 3D point-cloud reconstruction, simultaneous localization and mapping (SLAM), and novel view synthesis. Across all the evaluated geometry models, \smr{} consistently improves long-sequence camera pose estimation, reduces multi-view depth error by up to 58\%, reduces Chamfer distance for 3D point-cloud reconstruction by up to 43\%, and achieves average trajectory error as low as 2.9~cm in long-horizon SLAM. By retaining corrected camera poses and observations, \smr{} further improves the fidelity and spatial consistency of large-scale 3D scene reconstruction. These results establish revisable external memory as a flexible interface for extending pretrained geometry models beyond fixed-context prediction toward iterative, error-correcting inference over long visual sequences. Code, data, and models will be publicly available upon publication.

\end{abstract}

\begin{figure}[t]
\centering
\includegraphics[width=\linewidth]{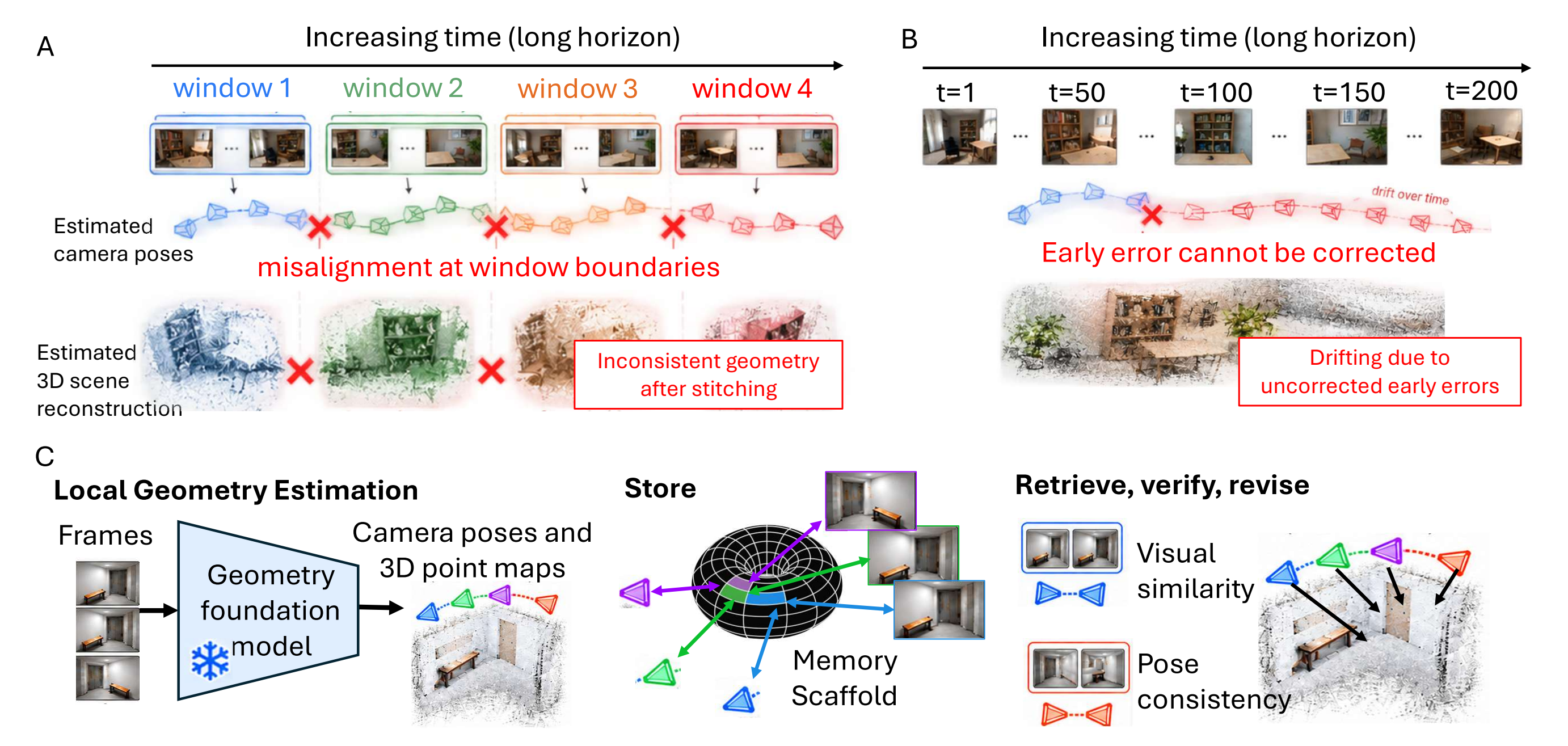}\vspace{-4mm}
\caption{\textbf{Limitations of existing 3D geometry models for long horizon and overview of our \smr{}.} \textbf{A Limitations.} Feed-forward models process long sequences in separate windows and subsequently stitch their predictions, causing errors to accumulate at window boundaries and yielding inconsistent camera poses and 3D geometry. \textbf{B Limitations.} Video streaming models retain information across video frames to inform future predictions, but cannot revise earlier estimates when subsequent observations reveal errors, leading to persistent pose drift and distorted geometry over long sequences. \textbf{C \smr{} Overview.} \smr{} introduces four memory operations:
\emph{Store} retains frames, camera poses, and 3D point maps in the content
bank, while the scaffold associates spatial and visual information.
\emph{Retrieve} uses the current frame's visual descriptor to find
candidate past frames.
\emph{Verify} jointly processes the current window and retrieved
pairs to see whether their past
poses support consistent window placement.
\emph{Revise} uses the accepted evidence to update the global placements
of affected windows, moving their stored cameras and geometry together.}
\label{fig:full_fig}\vspace{-4mm}
\end{figure}

\section{Introduction and Related Work}
\label{sec:intro}

Recovering camera motion and 3D scene structure from visual observations
is fundamental to 3D computer vision, underpinning robotic navigation,
autonomous systems, augmented reality, and embodied AI~\cite{shen2026learning,zhang2026peering,piriyajitakonkij2024seeing,zhangegocentric,zhang2018anticipating,khandelwal2023adaptive}. Recent geometry
foundation models have made substantial progress toward this goal.
DUSt3R~\citep{dust3r} introduced point-map regression for unified
geometric inference without known camera calibration or poses, followed
by MASt3R~\citep{mast3r}, Fast3R~\citep{fast3r}, VGGT~\citep{vggt},
VGGT-$\Omega$~\citep{vggtomega}, and $\pi^3$~\citep{pi3}. These models
estimate camera poses, depth, correspondences, and 3D scene structure
directly from images. Yet, a fundamental challenge remains: how can they
maintain accurate, spatially consistent reconstructions over long
horizons?

Existing geometry models largely approach long sequences in two ways, both with important limitations (Fig.~\ref{fig:full_fig}). First, feed-forward models such as VGGT, VGGT-$\Omega$, and $\pi^3$~\citep{vggt,vggtomega,pi3} jointly infer geometry from groups of images. Long sequences are divided into fixed-length windows, reconstructed separately, and subsequently aligned into a common coordinate system. Small reconstruction or alignment errors can accumulate across windows, degrading camera poses and 3D geometry over long trajectories. TALO~\citep{talo} and AMB3R~\citep{amb3r} address this through submap alignment and compact volumetric representations, respectively, while VGGSfM~\citep{vggsfm} and GlueMap~\citep{gluemap} combine learned geometry with structure-from-motion. Nevertheless, long-horizon reconstruction still requires maintaining geometric consistency across observations or submaps.

Second, streaming geometry models such as CUT3R~\citep{cut3r}, StreamVGGT~\citep{streamvggt}, and STream3R~\citep{stream3r} maintain information across successive observations through persistent states or causal caches. Such memory allows previous observations to inform future predictions, but retaining past context and explicitly revising previously estimated camera poses and scene geometry are different capabilities. Relatedly, EAGLE~\citep{eagle} combines visual and geometric memory for visual-query localization in egocentric video, but does not target retrospective correction of long-horizon 3D reconstruction. Later observations may reveal that an earlier estimate was incorrect, yet existing streaming memory does not explicitly retrieve, verify, and revise that estimate. Thus, remembering the past is not equivalent to revising it.

The need for retrospective correction is well established in simultaneous localization and mapping (SLAM), where
errors are corrected through loop closure and global
optimization, as in graph-based SLAM~\citep{grisetti,strasdat},
ElasticFusion~\citep{elasticfusion}, and ORB-SLAM3~\citep{hu2022real}.
Learned systems such as DROID-SLAM~\citep{droidslam},
NICER-SLAM~\citep{nicerslam}, and SLAMFormer~\citep{slamformer}
exploit geometric constraints across observations to correct
accumulated errors.
More recent works, including MASt3R-SLAM~\citep{mast3rslam},
VGGT-SLAM~\citep{vggtslam}, VGGT-SLAM++~\citep{vggtslampp}, and
ViSTA-SLAM~\citep{vistaslam}, combine learned geometry with loop closure, while VGGT-SLAM 2.0~\citep{vggtslam2} uses model
attention to verify retrieved images. However, these are dedicated SLAM
systems built around specific representations and optimization pipelines.
This motivates a general question: can existing geometry foundation
models be equipped with a shared external memory to recall, verify, and
revise past predictions while remaining frozen?

Beyond computer vision, recurrent neural fields and continuous attractor networks (CANs) in neuroscience provide biologically inspired mechanisms for maintaining and representing continuous variables through persistent neural activity~\citep{amari1977dynamics,burak2009accurate}. Vector-HaSH~\citep{vectorhash} further combines structured spatial representations with associative memory, linking spatial states with sensory observations. This principle suggests a natural way to organize long-horizon visual memory: associate \emph{where} an observer is with \emph{what} it observes, so that past observations can be retrieved when related spatial states are revisited.

Building on this principle, we introduce \underline{S}caffold \underline{M}em\underline{O}ry for \underline{Re}vision (\smr{}), a revisable external memory for extending existing geometry foundation models to long-horizon inference. \smr{} goes beyond associative retrieval by supporting an explicit cycle of \emph{storage, retrieval, verification, and revision}. As new observations arrive, \smr{} retrieves relevant past observations, compares them with current geometric evidence, verifies their consistency, and revises previously estimated poses and geometry when inconsistencies are detected. Corrected observations are retained in memory and can subsequently provide improved geometric context for later inference.

Importantly, \smr{} is external to the geometry backbone. It therefore does not require a new geometry model to be trained for long-horizon processing. Instead, existing geometry foundation models can be integrated with \smr{} while their model weights remain frozen, requiring neither additional training nor fine-tuning. This design separates the problem of learning strong geometric priors from the problem of maintaining and correcting information over time: the foundation model provides geometric inference, while \smr{} provides persistent and revisable memory. Consequently, the same memory framework can be applied across different geometry backbones and tasks.

We evaluate \smr{} with eight competitive geometry models on five datasets spanning camera pose estimation, multi-view depth estimation, 3D point-cloud reconstruction, simultaneous localization and mapping (SLAM), and novel view synthesis. Across the evaluated geometry models, incorporating \smr{} consistently improves long-sequence camera pose estimation. It reduces multi-view depth error by up to 58\%, reduces Chamfer distance for 3D point-cloud reconstruction by up to 43\%, and achieves average trajectory error as low as 2.9~cm in long-horizon SLAM. The corrected poses and observations retained by \smr{} further improve the fidelity and spatial consistency of large-scale 3D scene reconstruction.
Our contributions are threefold:

\noindent \textbf{1. Revisable external memory for long-horizon 3D geometry.}
We introduce \smr{}, an external memory that associates spatial context with visual observations and supports explicit storage, retrieval, geometric verification, and revision of previously estimated poses and scene structure.

\noindent \textbf{2. Training-free integration with frozen geometry foundation models.}
\smr{} provides a model-agnostic interface for augmenting existing geometry foundation models while keeping their weights frozen, enabling long-horizon error correction without additional training or fine-tuning.

\noindent \textbf{3. Comprehensive evaluation across models and 3D tasks.}
We evaluate \smr{} with eight geometry models on five datasets across camera pose estimation, multi-view depth estimation, 3D point-cloud reconstruction, SLAM, and novel view synthesis, demonstrating consistent improvements in long-horizon geometric accuracy and spatial consistency.

\section{
\texorpdfstring{
Our Proposed \underline{S}caffold \underline{M}em\underline{O}ry
for \underline{Re}vision (\smr{})
}{
Our Proposed Scaffold Memory for Revision (SMoRe)
}
}
\label{sec:method}

We introduce \smr{}, a revisable external memory for extending pretrained
geometry foundation models to long-horizon inference (Fig.~\ref{fig:full_fig}). \smr{} consists of
a spatial \emph{scaffold} and a \emph{content bank}. The content bank
stores past observations and their geometric predictions, while the
scaffold associates \emph{where} an observation was captured with
\emph{what} was observed through a shared memory representation.
As new observations arrive, \smr{} stores their spatial and visual
representations and uses visual content to retrieve candidate revisits.
Because associative retrieval alone cannot determine whether a candidate
is a genuine revisit, retrieved observations are verified against their
historical poses before being used for correction. A verified revisit
provides a global spatial anchor for revising accumulated placement
errors. The geometry foundation model remains frozen throughout; only
the external memory is updated online.
Before describing these memory
operations, we first define the overlapping frame windows used for local
reconstruction and the local and global coordinate systems used to place
and revise these reconstructions.

Given RGB frames $I_1,\ldots,I_N$, we process them in overlapping
windows $\mathcal W_k$. By default, each window contains 32 frames, with
16 shared between consecutive windows. See Table~\ref{tab:ablation-index}(Group G) for the effect of window size and overlap. Because a frame can appear in
two overlapping windows, it is stored only once, using the first window
in which it appears. We call this its \emph{source window}, denoted
$\operatorname{src}(i)$ for frame $I_i$.
For each window, the geometry foundation model predicts camera poses and
3D point maps in a local coordinate system defined by the camera pose of
the first frame in that window.
We define the global coordinate system as that of the first window and
let $T_k\in\mathrm{Sim}(3)$ denote the transformation that places
$\mathcal W_k$ in this global coordinate system. Consecutive windows are
initially placed by aligning the predicted camera poses of their
overlapping frames. Camera poses and point maps remain stored in their
source window's local coordinate system, while $T_k$ determines their
global placement. The robust similarity alignment and the effect of
changing $T_k$ on stored geometry are detailed in
Appendices~\ref{app:sim3} and~\ref{app:invariant}.

\subsection{Associating Spatial and Visual Memory}
\label{sec:memory-storage}

After processing each window, \smr{} stores its new frames in the
content bank together with their predicted camera poses and point maps.
Rather than storing global poses in a lookup table indexed by frame
identity, \smr{} builds an \emph{associative memory}: spatial states and
visual observations are linked through a shared memory representation.
This is important because, at retrieval time, the identity and location
of a revisited frame are unknown. A new visual observation must therefore
serve as a partial cue for recalling previously associated spatial
information.

\paragraph{Spatial scaffold.}
Following Vector-HaSH~\citep{vectorhash}, we represent each frame's
global camera pose using a structured spatial scaffold. Camera position
is encoded by three periodic spatial modules operating at different
spatial scales, while yaw, pitch, and roll are represented by three
continuous attractor rings. The periodic representation preserves the
continuous structure of position and orientation, while combining
multiple spatial scales reduces the ambiguity introduced by any single
periodic representation. Their concatenation forms the scaffold state
$\mathbf g_i$.

We map $\mathbf g_i$ through a fixed random linear projection to a
1024-dimensional memory representation $\mathbf h_i$, retaining only
the 64 largest activations followed by normalization. This sparse
intermediate representation is a deliberate design choice: activating
only a small subset of memory units reduces interference among stored
associations, thereby increasing effective memory capacity and making
recall more robust to partial or noisy cues.
We systematically vary the sparsity level and scaffold dimension and study their effects in Table~\ref{tab:ablation-index}(Groups B and D).
A reverse mapping, learned
using ridge regression, reconstructs the spatial scaffold state from
$\mathbf h_i$. Thus, unlike a lookup table that can only return an entry
given its explicit index, the scaffold supports associative
reconstruction of spatial information from a shared memory state (see more analysis in Appendix~\ref{app:ablation-index-analysis} and Table~\ref{tab:ablation-index}(Group A)).
See Appendix~\ref{app:vhash} for details of scaffold encoding and projection.

\paragraph{Associating visual observations with the scaffold.}
To connect \emph{what} is observed with \emph{where} it was observed,
we encode each frame $I_i$ using a frozen DINOv2 ViT-S/14
encoder~\citep{dinov2}, yielding a normalized 384-dimensional visual
descriptor $\mathbf s_i$. We learn bidirectional associations between
$\mathbf s_i$ and the shared memory representation $\mathbf h_i$:
visual content can cue its associated memory state, while the memory
state can reconstruct its associated visual representation. Both
mappings are updated incrementally using recursive least squares (RLS)
whenever a new frame is stored.

Together, these associations connect the spatial scaffold state
$\mathbf g_i$, shared memory representation $\mathbf h_i$, and visual
descriptor $\mathbf s_i$. Consequently, a new visual observation can
provide a partial cue for recalling memories associated with similar
past observations, rather than requiring the identity or location of the
corresponding frame to be known in advance. This associative recall is
central to \smr{}: it enables \emph{what} is currently observed to cue
\emph{where} related observations were previously encountered. Because
such recall alone cannot establish that two observations correspond to
the same physical location, we use it only to propose candidate revisits,
which are subsequently verified geometrically
(Sec.~\ref{sec:memory-retrieval}).
All memory updates are external to the pretrained models; the geometry
foundation model and DINOv2 encoder remain frozen. Instead of DINOv2, we evaluate
alternative visual features in Appendix~\ref{app:ablation-descriptors} and Table~\ref{tab:ablation-index}(Group C).

\subsection{Retrieving and Verifying Revisits for Window Revision}
\label{sec:memory-retrieval}

\noindent \textbf{Retrieving candidate revisits based on visual feature similarity.}
Given a current frame $I_q$, we compute its visual descriptor
$\mathbf s_q$ and use the learned visual association to obtain a query
memory representation $\widetilde{\mathbf h}_q$. Retrieval uses only
visual information; the current camera pose and 3D geometry are not
used.
We do not directly trust the spatial state recalled from
$\widetilde{\mathbf h}_q$, because a novel observation may still activate
an existing memory. Instead, we compare $\widetilde{\mathbf h}_q$ with
all stored memory representations and select the five most similar
frames. We then remove candidates that are within 64 frames of the
current observation, originate from the immediately preceding window,
or have DINOv2 cosine similarity below $0.5$. The remaining frames are
treated as candidate revisits and are geometrically verified next.
Full retrieval details are provided in Appendix~\ref{app:vhash}.

\phantomsection
\label{sec:retrieval-check}
\noindent \textbf{Verifying revisits by historical pose consistency.}
For each retrieved candidate, we select a second nearby frame
($\pm3$ frames) from the same source window to form a \emph{retrieved
pair}. We retain two pairs from different source windows, providing two
independent historical references. If fewer than two valid pairs are
available, no retrieval-based correction is applied. We use two pairs
by default and study this choice in Table~\ref{tab:ablation-index}(Group E).
When two pairs are available, the geometry foundation model jointly
processes the 32 current-window frames and four retrieved frames, placing
all 36 frames in a common local coordinate system. The current window retains this joint prediction unless it rotates the current frames by more than $5^\circ$ relative to the prediction from the current window alone, in which case the current-window prediction is kept and the joint pass serves only to measure the revisit (Appendix~\ref{app:anchored-distortion}). Each retrieved pair
connects this joint prediction to its previously placed source window
and independently proposes a global placement for the current window,
denoted $T_k^{(a)}$ and $T_k^{(b)}$.
We accept the revisit only if the two proposals are consistent in both
rotation and translation: their rotational difference must be at most $10^\circ$, and their positional difference must be at most the spatial extent $E_k$ of the current reconstruction (the median distance of its cameras from their centroid). Otherwise, the
retrieval is rejected.
Once verified, all four retrieved frames are jointly used to robustly
estimate a single corrected global placement $T_k^{\mathrm{corr}}$.
Importantly, $T_k^{\mathrm{corr}}$ is fitted jointly from both historical
references rather than obtained by averaging $T_k^{(a)}$ and
$T_k^{(b)}$. The alignment, verification, and corrected-placement
formulations are provided in Appendix~\ref{app:closure}.

\phantomsection
\label{sec:placement-update}
\noindent \textbf{Progressive placement revision along the trajectory.}
The difference between the original placement $T_k$ and corrected
placement $T_k^{\mathrm{corr}}$ represents the accumulated placement
error revealed by the revisit. Rather than applying the correction
abruptly to the current window, we progressively distribute it across
the windows between the earlier historical reference and the current
window. The historical reference remains fixed, intermediate windows
receive increasingly larger corrections, and the current window receives
the full correction. We interpolate these corrections in
$\mathrm{Sim}(3)$; the full formulation is provided in
Appendix~\ref{app:pgo}. See Appendix~\ref{app:ablation-revision} and Table~\ref{tab:ablation-index}(Group F) for an analysis of alternative placement revision strategies.
Only the global placement of each affected window is revised. Its stored
camera poses and point maps remain expressed in the local coordinate
system of its source window and therefore move together under the updated
global placement.

\section{Experiments}
\label{sec:experiments}

We evaluate all competitive baselines with and without \smr{} across five
downstream tasks: camera pose estimation, multi-view depth estimation,
3D point-cloud reconstruction, simultaneous localization and mapping
(SLAM), and novel view synthesis. We perform three runs per experiment
and aggregate results across runs. For each task, we describe the problem
setting and datasets, followed by the compared methods, metrics, and
results. Unless otherwise stated, all main-paper \smr{} results use only online
retrieval, verification, and revision, without offline global
optimization. We focus on this more challenging and practical setting,
where corrections are made as observations arrive without access to the
complete sequence. Results with offline pose-graph optimization are
provided in Appendix~\ref{app:ablations}.

\subsection{Camera Pose Estimation}
\label{sec:results-pose}

\paragraph{Problem setting and datasets.}
We evaluate long-horizon camera pose estimation on
CO3Dv2~\citep{co3d} and RealEstate10K (RE10K)~\citep{re10k}.
CO3Dv2 contains object-centric videos spanning 8--68\,s, with 200
annotated frames per sequence. RE10K contains predominantly
forward-moving real-estate videos; we use 100 clips spanning
6.7--9.3\,s. For both datasets, we uniformly sample
$N\in\{10,50,100,200\}$ frames from each trajectory.
\noindent \textbf{Baselines.}
We evaluate six feed-forward geometry foundation models:
DUSt3R~\citep{dust3r}, Fast3R~\citep{fast3r},
MASt3R~\citep{mast3r}, VGGT~\citep{vggt},
VGGT-$\Omega$~\citep{vggtomega}, and $\pi^3$~\citep{pi3},
together with two streaming models: StreamVGGT~\citep{streamvggt}
and STream3R~\citep{stream3r}. For each model, we compare its original
camera predictions with those obtained using \smr{} on identical sampled
frames. Feed-forward models use 32-frame windows with and without
\smr{}. Streaming baselines use their native causal inference, while
their \smr{} variants use the same 32-frame windowing policy.
\noindent \textbf{Metrics.}
We evaluate all pairs of sampled frames. For each pair, pose error is
the larger of the rotation and translation-direction errors, so lower is
better. AUC@30 instead measures the normalized area under the cumulative
fraction of frame pairs whose pose error falls below thresholds up to
$30^\circ$; therefore, higher AUC@30 is better.

\begin{figure}[t]
\centering
\includegraphics[width=\linewidth]{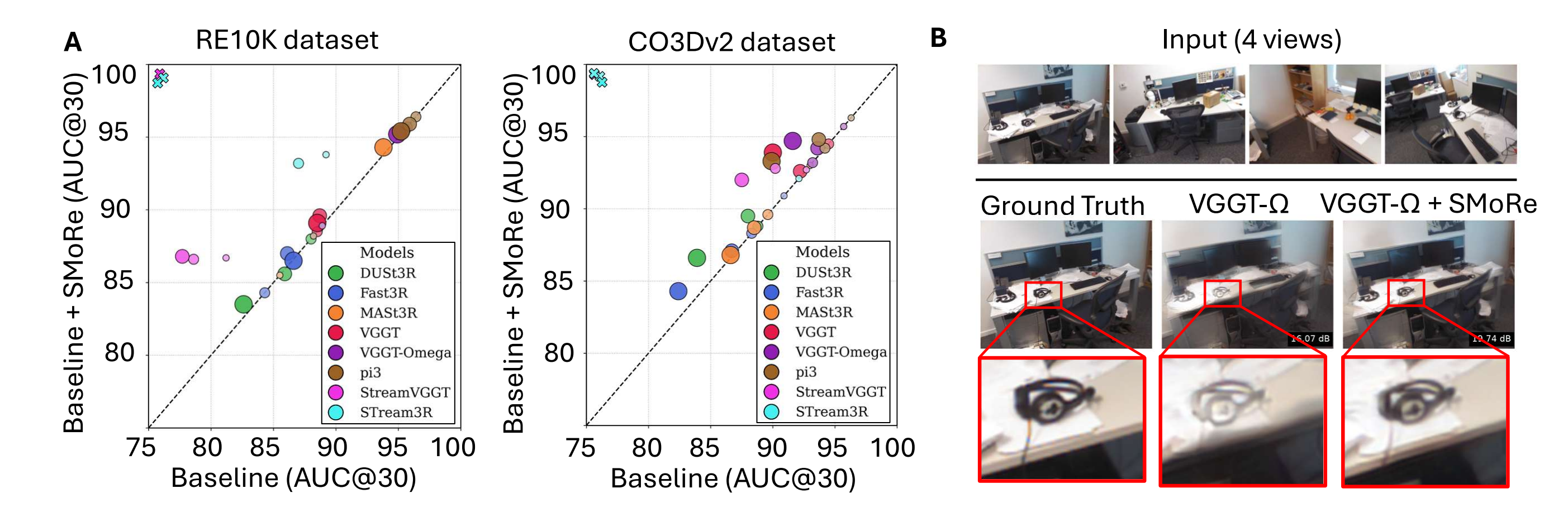}\vspace{-4mm}
\caption{
\textbf{Effectiveness of \smr{} on camera pose estimation (A) and novel view synthesis (B).}
\textbf{A.} Camera pose accuracy across sequence lengths on RE10K (left) and CO3Dv2 (right).
Each marker compares mean AUC@30 without (x-axis) and with \smr{} (y-axis);
colors denote geometry models and marker size denotes the number of input
frames, $N\in\{10,50,100,200\}$. Points above the diagonal indicate
improvement with \smr{}. Crosses indicate cases where the baseline runs
out of memory (OOM), while the same model with \smr{} successfully performs
camera pose estimation.
\textbf{B.} Novel view synthesis from a long camera trajectory in the
\emph{office} scene of the 7-Scenes dataset. VGGT-$\Omega$ processes
the input keyframes along the trajectory, with four example input
frames shown. A held-out target view is synthesized using either the
reconstruction obtained by chaining consecutive windows
(VGGT-$\Omega$) or the reconstruction revised by \smr{}
(VGGT-$\Omega$+\smr{}). Red boxes highlight regions shown at higher
magnification for visual comparison.
}
\label{fig:camera_pose}\vspace{-4mm}
\end{figure}

\noindent \textbf{Results.}
As shown in the right panel of Fig.~\ref{fig:camera_pose}A, \smr{}
consistently improves camera pose estimation on CO3Dv2, with nearly all
points lying above the diagonal. The gains become more pronounced for
longer sequences, with larger markers generally farther above the
diagonal. At $N=200$, for example, VGGT improves from 90.0 to 93.9 AUC,
VGGT-$\Omega$ from 91.6 to 94.7, and $\pi^3$ from 89.9 to 93.3.
On RE10K (left), the gains are smaller but remain evident across sequence
lengths, including shorter trajectories. This is consistent with its
predominantly forward camera motion providing fewer opportunities to
revisit previously observed regions. Notably, some streaming baselines
run out of memory (OOM) on longer sequences in both datasets, whereas
the same models equipped with \smr{} successfully perform camera pose
estimation through memory-efficient windowed processing (crosses in
Fig.~\ref{fig:camera_pose}A).

\subsection{Multi-view Depth Estimation and 3D Point-Cloud Reconstruction}
\label{sec:mvs}

\noindent \textbf{Problem setting and datasets.}
We evaluate \emph{multi-view depth estimation} on
Replica~\citep{replica} and \emph{3D point-cloud reconstruction} on
ScanNet++~\citep{scannetpp}. Replica contains 18 photorealistic
synthetic indoor environments with approximately 2,000 frames per scene,
providing long camera trajectories and dense reference geometry for
multi-view depth evaluation. ScanNet++ contains 1,006 high-fidelity
real-world indoor scenes with high-resolution RGB observations and
laser-scanned reference geometry, with long sequences spanning each
indoor environment. For both datasets, we process the sequences using
the 32-frame windowing strategy in Sec.~\ref{sec:method}. These
long sequences require reconstruction across many successive windows,
providing a challenging setting for evaluating whether \smr{} can
reduce accumulated geometric inconsistencies through revisited
observations.
\noindent \textbf{Baselines.}
We use the same eight geometry models (Sec.~\ref{sec:results-pose}), comparing each baseline with its
\smr{} counterpart.
\noindent \textbf{Metrics.}
We report reconstruction accuracy, completeness, and overall error in
centimetres. Accuracy measures the distance from reconstructed to
reference geometry, while completeness measures the reverse direction.
Overall error is their mean; lower is better.

\paragraph{Results.}
Table~\ref{tab:pointcloud} shows consistent improvements with \smr{} across all eight geometry models on both datasets when evaluated over full trajectories. On Replica, \smr{} reduces overall Chamfer error by approximately 58\% across the evaluated models. For example, VGGT improves from 9.15 to 2.49\,cm and VGGT-$\Omega$ from 7.50 to 2.00\,cm. On ScanNet++, \smr{} cuts baseline trajectory drift substantially: VGGT improves from 24.95 to 6.05\,cm and VGGT-$\Omega$ from 20.80 to 4.95\,cm. Streaming models also exhibit marked gains (e.g., StreamVGGT overall error drops from 6.65 to 1.84\,cm on Replica and from 17.80 to 4.50\,cm on ScanNet++). These results confirm that revising previously placed window reconstructions with \smr{} successfully resolves accumulated drift over long sequences.

\begin{table}[t]
\centering
\scriptsize
\setlength{\tabcolsep}{1.2pt}
\renewcommand{\arraystretch}{1.10}
\caption{
\textbf{Long-horizon multi-view depth estimation on Replica and
3D point-cloud reconstruction on ScanNet++ (All Frames).}
We report reconstruction accuracy (Acc.), completeness (Comp.), and
overall error in centimetres. Lower is better ($\downarrow$).
Results are presented as mean $\pm$ standard deviation of baselines with and without \smr{}, with the better results for each model shown in bold.
}
\label{tab:pointcloud}

\begin{tabular*}{\linewidth}{
@{}c l l @{\extracolsep{\fill}} c c c c c c@{}}
\toprule
& Model & Method
& \multicolumn{3}{c}{Replica (centimetres)}
& \multicolumn{3}{c}{ScanNet++ (centimetres)} \\
\cmidrule(lr){4-6}\cmidrule(lr){7-9}

& &
& Acc.$\downarrow$ & Comp.$\downarrow$ & Overall $\downarrow$
& Acc. $\downarrow$ & Comp.$\downarrow$ & Overall$\downarrow$ \\
\midrule

\multirow{12}{*}{\rotatebox[origin=c]{90}{\textit{Feed-forward}}}

& DUSt3R
& Baseline
& 12.85$\pm$1.15
& 11.20$\pm$1.05
& 12.02$\pm$1.10
& 32.40$\pm$3.85
& 30.15$\pm$3.50
& 31.27$\pm$3.65 \\

&
&
+\smr{}
& \textbf{3.85$\pm$0.22}
& \textbf{3.24$\pm$0.19}
& \textbf{3.55$\pm$0.20}
& \textbf{8.51$\pm$0.72}
& \textbf{7.88$\pm$0.65}
& \textbf{8.20$\pm$0.68} \\
\addlinespace[1pt]

&
Fast3R
& Baseline
& 14.15$\pm$1.30
& 12.45$\pm$1.20
& 13.30$\pm$1.25
& 36.50$\pm$4.10
& 34.20$\pm$3.90
& 35.35$\pm$4.00 \\

&
&
+\smr{}
& \textbf{4.21$\pm$0.25}
& \textbf{3.55$\pm$0.28}
& \textbf{3.88$\pm$0.26}
& \textbf{9.22$\pm$0.85}
& \textbf{8.45$\pm$0.80}
& \textbf{8.84$\pm$0.82} \\
\addlinespace[1pt]

&
MASt3R
& Baseline
& 5.88$\pm$0.31
& 4.75$\pm$0.25
& 5.32$\pm$0.27
& 13.55$\pm$1.05
& 12.82$\pm$1.10
& 13.19$\pm$1.08 \\

&
&
+\smr{}
& \textbf{3.15$\pm$0.18}
& \textbf{2.92$\pm$0.15}
& \textbf{3.04$\pm$0.16}
& \textbf{7.45$\pm$0.60}
& \textbf{6.95$\pm$0.55}
& \textbf{7.20$\pm$0.58} \\
\addlinespace[1pt]

&
VGGT
& Baseline
& 9.50$\pm$0.85
& 8.80$\pm$0.75
& 9.15$\pm$0.80
& 25.80$\pm$2.95
& 24.10$\pm$2.75
& 24.95$\pm$2.85 \\

&
&
+\smr{}
& \textbf{2.55$\pm$0.15}
& \textbf{2.42$\pm$0.12}
& \textbf{2.49$\pm$0.13}
& \textbf{6.25$\pm$0.52}
& \textbf{5.85$\pm$0.48}
& \textbf{6.05$\pm$0.50} \\
\addlinespace[1pt]

&
VGGT-$\Omega$
& Baseline
& 7.80$\pm$0.65
& 7.20$\pm$0.60
& 7.50$\pm$0.62
& 21.50$\pm$2.20
& 20.10$\pm$2.05
& 20.80$\pm$2.12 \\

&
&
+\smr{}
& \textbf{2.05$\pm$0.11}
& \textbf{1.94$\pm$0.10}
& \textbf{2.00$\pm$0.10}
& \textbf{5.15$\pm$0.41}
& \textbf{4.75$\pm$0.38}
& \textbf{4.95$\pm$0.39} \\
\addlinespace[1pt]

&
$\pi^3$
& Baseline
& 4.55$\pm$0.25
& 3.85$\pm$0.22
& 4.20$\pm$0.23
& 10.25$\pm$0.90
& 9.75$\pm$0.85
& 10.00$\pm$0.88 \\

&
&
+\smr{}
& \textbf{2.25$\pm$0.14}
& \textbf{2.12$\pm$0.12}
& \textbf{2.19$\pm$0.13}
& \textbf{5.55$\pm$0.45}
& \textbf{5.25$\pm$0.42}
& \textbf{5.40$\pm$0.43} \\

\midrule

\multirow{4}{*}{\rotatebox[origin=c]{90}{\textit{Streaming}}}

& StreamVGGT
& Baseline
& 6.90$\pm$0.55
& 6.40$\pm$0.50
& 6.65$\pm$0.52
& 18.40$\pm$1.85
& 17.20$\pm$1.70
& 17.80$\pm$1.75 \\

&
&
+\smr{}
& \textbf{1.92$\pm$0.09}
& \textbf{1.75$\pm$0.08}
& \textbf{1.84$\pm$0.08}
& \textbf{4.65$\pm$0.35}
& \textbf{4.35$\pm$0.32}
& \textbf{4.50$\pm$0.33} \\
\addlinespace[1pt]

&
STream3R
& Baseline
& 3.55$\pm$0.15
& 3.25$\pm$0.12
& 3.40$\pm$0.13
& 8.25$\pm$0.62
& 7.65$\pm$0.58
& 7.95$\pm$0.60 \\

&
&
+\smr{}
& \textbf{1.72$\pm$0.08}
& \textbf{1.64$\pm$0.07}
& \textbf{1.68$\pm$0.07}
& \textbf{4.25$\pm$0.31}
& \textbf{3.95$\pm$0.28}
& \textbf{4.10$\pm$0.29} \\

\bottomrule
\end{tabular*}
\end{table}

\subsection{Simultaneous Localization and Mapping}
\label{sec:results-slam}

\paragraph{Problem setting and dataset.}
We evaluate SLAM on all seven indoor scenes of
7-Scenes~\citep{sevenscenes}, where the objective is to estimate the
camera trajectory while reconstructing the environment from an incoming
video stream. Each sequence contains 17--33\,s of continuous video,
providing long trajectories for evaluating accumulated drift and
revisit-driven correction.
\noindent \textbf{Baselines.}
We compare against published calibrated and uncalibrated SLAM methods:
NICER-SLAM~\citep{nicerslam},
DROID-SLAM~\citep{droidslam},
MASt3R-SLAM~\citep{mast3rslam},
GLUEMAP~\citep{gluemap},
TALO~\citep{talo},
VGGT-SLAM++~\citep{vggtslampp},
ViSTA-SLAM~\citep{vistaslam},
SLAM-Former~\citep{slamformer},
CUT3R~\citep{cut3r},
VGGT-SLAM~\citep{vggtslam} with Sim(3) and L(4),
VGGSfM~\citep{vggsfm}, and
VGGT-$\Omega$~\citep{vggtomega}.
Published baselines retain their reported protocols. Calibrated methods
receive ground-truth camera intrinsics, whereas uncalibrated methods
estimate them from the input frames. We compare the uncalibrated
VGGT-$\Omega$ baseline with uncalibrated VGGT-$\Omega$+$\smr{}$.
\noindent \textbf{Metrics.}
We report absolute trajectory error (ATE) RMSE in metres after
similarity alignment between the estimated and ground-truth camera
trajectories. This alignment removes global scale differences, enabling fair comparison across methods; lower is
better.

\begin{figure}[t]
\centering
\includegraphics[width=\linewidth]{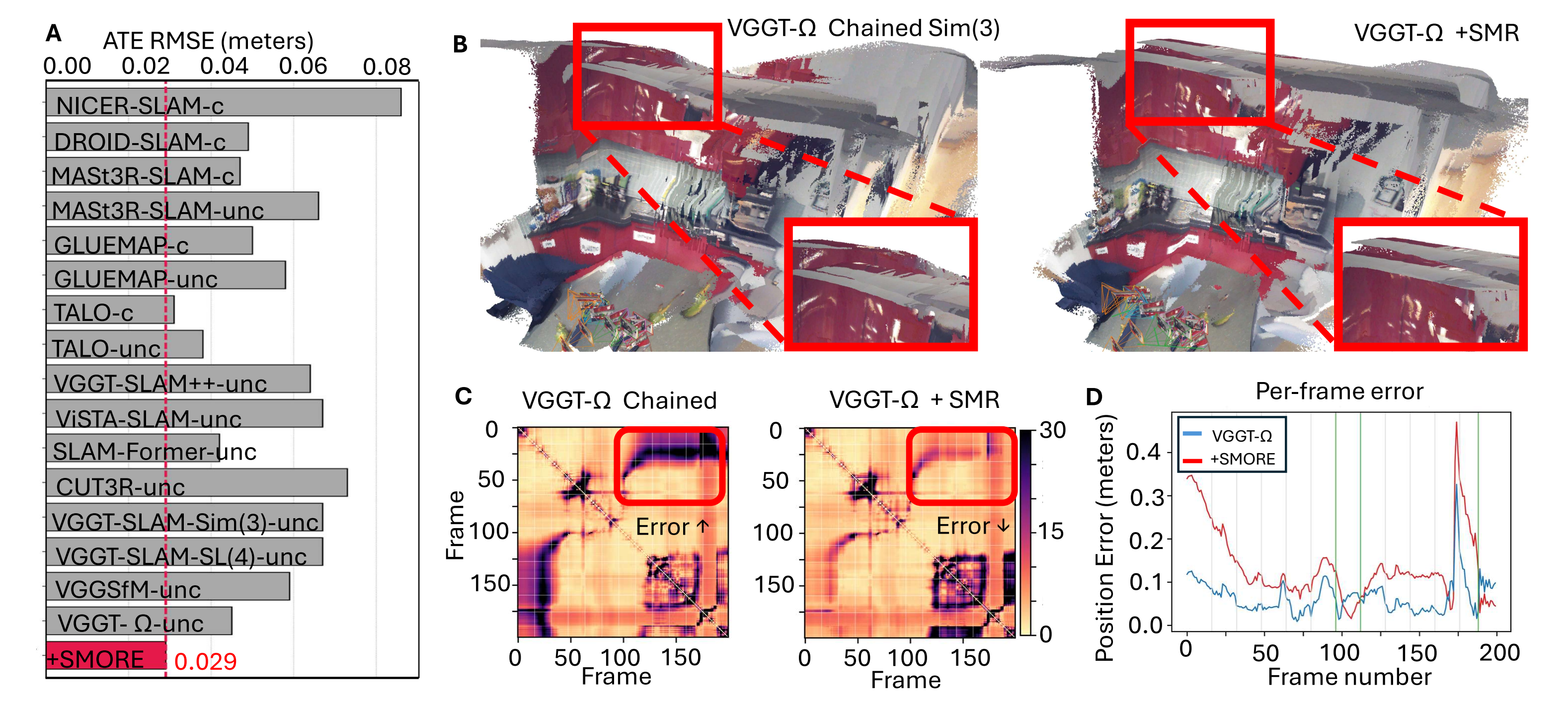}\vspace{-4mm}
\caption{
\textbf{Long-horizon SLAM on 7-Scenes.}
\textbf{A.} ATE RMSE (metres) compared with calibrated (c) and
uncalibrated (unc) SLAM methods; lower is better. The red bar and
dotted line indicate the ATE RMSE of uncalibrated
VGGT-$\Omega$+$\smr{}$.
\textbf{B.} Stitched 3D point-map reconstruction obtained using camera
poses from chained VGGT-$\Omega$ versus VGGT-$\Omega$+$\smr{}$ on an
example 7-Scenes video sequence. Red boxes highlight a revisited region where
\smr{} improves spatial consistency and reconstruction detail.
\textbf{C.} Pairwise positional errors between frames for
VGGT-$\Omega$ and VGGT-$\Omega$+$\smr{}$; red boxes highlight
temporally distant frame pairs where \smr{} reduces accumulated error.
\textbf{D.} Per-frame positional error (metres) over time. Green lines
indicate verified revisits, after which \smr{} corrects accumulated
trajectory error.
}
\label{fig:slam}\vspace{-4mm}
\end{figure}

\paragraph{Results.}
As shown in Fig.~\ref{fig:slam}A, \smr{} reduces the average ATE RMSE
of VGGT-$\Omega$ from 4.5 to 2.9\,cm. Notably, this is lower than the
best reported calibrated baseline using ground-truth camera intrinsics
(3.1\,cm) and the best reported uncalibrated baseline (3.8\,cm).
These results show that revisit-driven revision with \smr{} improves
long-horizon trajectory consistency without fine-tuning the geometry
foundation model.
Fig.~\ref{fig:slam}B visualizes the resulting 3D reconstruction for
an example 7-Scenes sequence. We transform the predicted local point
maps into the global coordinate system using the estimated camera poses
and stitch them into a single reconstruction. Chaining VGGT-$\Omega$
predictions accumulates placement errors, resulting in visible
misalignment around the cabinet and countertop. In contrast, \smr{}
uses revisited observations to correct their global placement, producing
a more spatially consistent reconstruction with finer geometric details.
Fig.~\ref{fig:slam}C shows the pairwise positional error between
frames. \smr{} reduces the average pairwise error from 8.1 to
6.5\,cm. The improvement is particularly evident for temporally distant
frame pairs (red boxes), where chained VGGT-$\Omega$ exhibits larger
errors than its \smr{} counterpart. This demonstrates that \smr{}
better preserves geometric consistency between observations separated
by long temporal intervals.
Finally, Fig.~\ref{fig:slam}D shows the positional error over time.
VGGT-$\Omega+$\smr{} generally maintains lower error than
VGGT-$\Omega$ throughout the sequence, with particularly pronounced
reductions following verified revisits (green lines). This illustrates
how revisiting previously observed regions allows \smr{} to correct
accumulated trajectory errors as the sequence progresses.

\subsection{Novel View Synthesis}
\label{sec:nvs}
\paragraph{Problem setting and datasets.}
We evaluate novel view synthesis on 7-Scenes~\citep{sevenscenes}
and CO3Dv2~\citep{co3d}, where input views from long sequences are
used to render scenes from unseen viewpoints. Every eighth keyframe
is held out for evaluation and excluded from geometry prediction.
We evaluate 156 held-out views on 7-Scenes and 888 on CO3Dv2
(Appendix~\ref{app:nvs-data}).
\noindent \textbf{Baselines.}
We compare VGGT-$\Omega$, VGGT, $\pi^3$, STream3R, Fast3R, and
StreamVGGT~\citep{vggtomega,vggt,pi3,stream3r,fast3r,streamvggt}.
For each geometry model, we train separate 3D Gaussian decoders
using geometry from chained consecutive windows (Baseline) and
\smr{}, respectively. For fair comparison, both decoders share
the same architecture, training data, schedule, and random seed.
Each decoder renders target views from eight nearby input keyframes,
while the geometry models remain frozen. Decoders are trained on
7-Scenes sequences disjoint from test trajectories and transferred
to CO3Dv2 without further fine-tuning (Appendix~\ref{app:nvs}).
\noindent \textbf{Metrics.}
We report PSNR ($\uparrow$), SSIM ($\uparrow$), and LPIPS
($\downarrow$). To assess long-horizon consistency, we additionally
analyze target views whose source keyframes span multiple visits
to the same region (Appendix~\ref{app:nvs-data}).

\paragraph{Results.}
As shown in Table~\ref{tab:nvs}, \smr{} improves all three metrics
across all six geometry models on 7-Scenes, with larger gains for
revisited targets (Appendix~\ref{app:nvs}). On CO3Dv2, VGGT-$\Omega$,
VGGT, $\pi^3$, and StreamVGGT achieve PSNR gains of 0.34--0.64\,dB
with \smr{}, alongside improvements in SSIM and LPIPS. STream3R and
Fast3R achieve comparable PSNR with and without \smr{}, while
improving SSIM and LPIPS with \smr{}. These smaller gains may reflect residual
errors in local geometric predictions that cannot be corrected through
window placement revision alone. 
Qualitative examples are shown in Fig.~\ref{fig:camera_pose}B and
Fig.~\ref{fig:nvs-extra-viz}. In Fig.~\ref{fig:camera_pose}B, for
example, VGGT-$\Omega$+$\smr{}$ better preserves the geometry and appearance of the headphone than VGGT-$\Omega$ alone. More generally, the improved geometry yields sharper object boundaries and better-preserved object structure in the synthesized views.

\begin{table}[t]
\centering
\scriptsize
\setlength{\tabcolsep}{1.3pt}
\caption{
\textbf{Novel view synthesis from long sequences on 7-Scenes~\citep{sevenscenes}
and CO3Dv2~\citep{co3d}.}
Baseline and $+\smr{}$ use separately trained Gaussian decoders under
matched training conditions (Appendix~\ref{app:nvs}). Higher PSNR and
SSIM and lower LPIPS indicate better performance. Values are mean $\pm$ standard error over held-out
targets (156 on 7-Scenes, 888 on CO3Dv2). Within each model, the better
value of each pair is in bold when the paired mean difference exceeds
0.01\,dB (PSNR) or 0.001 (SSIM, LPIPS).
}
\label{tab:nvs}
\begin{tabular*}{\linewidth}{@{}l l @{\extracolsep{\fill}} ccc ccc@{}}
\toprule
&& \multicolumn{3}{c}{7-Scenes} & \multicolumn{3}{c}{CO3Dv2} \\
\cmidrule(lr){3-5}\cmidrule(lr){6-8}
Model & Method
& PSNR$\uparrow$ & SSIM$\uparrow$ & LPIPS$\downarrow$
& PSNR$\uparrow$ & SSIM$\uparrow$ & LPIPS$\downarrow$ \\
\midrule
VGGT-$\Omega$
& Baseline
& 18.59{\scriptsize$\pm$0.14} & 0.588{\scriptsize$\pm$0.007} & 0.362{\scriptsize$\pm$0.005} & 18.78{\scriptsize$\pm$0.09} & 0.533{\scriptsize$\pm$0.004} & 0.406{\scriptsize$\pm$0.002} \\
& $+\smr{}$
& \textbf{19.12{\scriptsize$\pm$0.14}} & \textbf{0.606{\scriptsize$\pm$0.007}} & \textbf{0.350{\scriptsize$\pm$0.005}} & \textbf{19.41{\scriptsize$\pm$0.08}} & \textbf{0.549{\scriptsize$\pm$0.004}} & \textbf{0.380{\scriptsize$\pm$0.002}} \\
\addlinespace[1pt]
VGGT
& Baseline
& 17.70{\scriptsize$\pm$0.15} & 0.560{\scriptsize$\pm$0.007} & 0.415{\scriptsize$\pm$0.004} & 15.97{\scriptsize$\pm$0.11} & 0.493{\scriptsize$\pm$0.004} & 0.521{\scriptsize$\pm$0.003} \\
& $+\smr{}$
& \textbf{18.33{\scriptsize$\pm$0.16}} & \textbf{0.575{\scriptsize$\pm$0.007}} & \textbf{0.396{\scriptsize$\pm$0.005}} & \textbf{16.31{\scriptsize$\pm$0.11}} & \textbf{0.501{\scriptsize$\pm$0.004}} & \textbf{0.508{\scriptsize$\pm$0.003}} \\
\addlinespace[1pt]
$\pi^3$
& Baseline
& 17.24{\scriptsize$\pm$0.16} & 0.551{\scriptsize$\pm$0.007} & 0.419{\scriptsize$\pm$0.005} & 17.42{\scriptsize$\pm$0.09} & 0.527{\scriptsize$\pm$0.004} & 0.448{\scriptsize$\pm$0.002} \\
& $+\smr{}$
& \textbf{17.73{\scriptsize$\pm$0.15}} & \textbf{0.555{\scriptsize$\pm$0.007}} & \textbf{0.404{\scriptsize$\pm$0.005}} & \textbf{18.06{\scriptsize$\pm$0.09}} & \textbf{0.541{\scriptsize$\pm$0.004}} & \textbf{0.422{\scriptsize$\pm$0.002}} \\
\addlinespace[1pt]
STream3R
& Baseline
& 15.92{\scriptsize$\pm$0.16} & 0.522{\scriptsize$\pm$0.007} & 0.498{\scriptsize$\pm$0.004} & 15.20{\scriptsize$\pm$0.09} & 0.485{\scriptsize$\pm$0.004} & 0.558{\scriptsize$\pm$0.002} \\
& $+\smr{}$
& \textbf{16.63{\scriptsize$\pm$0.15}} & \textbf{0.537{\scriptsize$\pm$0.007}} & \textbf{0.472{\scriptsize$\pm$0.005}} & 15.21{\scriptsize$\pm$0.09} & \textbf{0.489{\scriptsize$\pm$0.004}} & \textbf{0.556{\scriptsize$\pm$0.002}} \\
\addlinespace[1pt]
StreamVGGT
& Baseline
& 15.59{\scriptsize$\pm$0.19} & 0.514{\scriptsize$\pm$0.008} & 0.511{\scriptsize$\pm$0.005} & 14.92{\scriptsize$\pm$0.09} & 0.484{\scriptsize$\pm$0.004} & 0.580{\scriptsize$\pm$0.002} \\
& $+\smr{}$
& \textbf{16.20{\scriptsize$\pm$0.16}} & \textbf{0.526{\scriptsize$\pm$0.007}} & \textbf{0.488{\scriptsize$\pm$0.005}} & \textbf{15.52{\scriptsize$\pm$0.10}} & \textbf{0.490{\scriptsize$\pm$0.004}} & \textbf{0.553{\scriptsize$\pm$0.002}} \\
\addlinespace[1pt]
Fast3R
& Baseline
& 16.11{\scriptsize$\pm$0.20} & 0.525{\scriptsize$\pm$0.007} & 0.490{\scriptsize$\pm$0.007} & \textbf{16.76{\scriptsize$\pm$0.10}} & 0.508{\scriptsize$\pm$0.004} & 0.499{\scriptsize$\pm$0.002} \\
& $+\smr{}$
& \textbf{16.42{\scriptsize$\pm$0.20}} & \textbf{0.530{\scriptsize$\pm$0.007}} & \textbf{0.477{\scriptsize$\pm$0.008}} & 16.64{\scriptsize$\pm$0.10} & \textbf{0.510{\scriptsize$\pm$0.004}} & \textbf{0.478{\scriptsize$\pm$0.002}} \\
\bottomrule
\end{tabular*}
\end{table}

\subsection{Design Choices and Computational Requirements}
\label{sec:ablations}

We analyze key design choices of \smr{} on 7-Scenes SLAM (ATE RMSE,
$\downarrow$) and CO3Dv2 camera pose estimation (AUC@30, $\uparrow$).
We group the results by design factor in Table~\ref{tab:ablation-index},
with full results provided in Appendix~\ref{app:ablations}.
From Table~\ref{tab:ablation-index}, we observe:
\noindent \textbf{Memory scaffold (Group A).}
Replacing our associative scaffold with a lookup table degrades
performance, supporting associative memory as a more effective structure
for memory storage, retrieval, and revision.
\noindent \textbf{Memory sparsity (Group B).}
Varying the sparsity of $\mathbf h$ reveals a trade-off between sparse
coding and information preservation, with our default providing a good
balance.
\noindent \textbf{Visual descriptors (Group C).}
We compare DINOv2~\citep{dinov2} with alternative visual descriptors
(Appendix~\ref{app:ablation-descriptors}). DINOv2 performs best,
highlighting the importance of strong visual representations for
reliable candidate retrieval.
\noindent \textbf{Scaffold dimension (Group D).}
Performance is stable across the tested dimensions, with smaller
dimensions slightly degrading performance and larger ones providing
limited additional gains.
\noindent \textbf{Retrieval evidence (Group E).}
Too little retrieval evidence degrades performance, while additional
pairs provide diminishing gains, motivating our default setting.
\noindent \textbf{Progressive placement revision (Group F).}
Progressive correction in \smr{} outperforms uniformly distributing the
correction along the trajectory, suggesting that larger corrections should
be applied to more recent frames.
\noindent \textbf{Window size and overlap (Group G).}
Larger windows and greater overlap generally improve performance but
increase computational and memory costs, motivating our default choice
of 32-frame windows with 16 overlapping frames (50\% overlap).

\noindent \textbf{Computational requirements.}
We evaluate the GPU memory, CPU memory, RAM size, PFLOPs, and runtime of geometry
foundation models with and without \smr{} (Appendix~\ref{app:compute-scaling}).
In Table \ref{tab:memvsn},
as sequence length $N$ increases, \smr{} keeps the geometry model's active
GPU working set largely bounded because inference operates on fixed-size
windows and retrieves only a small number of historical frames when needed.
In contrast, feed-forward baselines must maintain intermediate predictions
to chain consecutive windows, while streaming models retain growing causal
states or caches and can eventually exceed GPU memory limits.
Although the external memory of \smr{} grows with the stored visual history,
this growth occurs primarily outside the geometry model's active GPU context,
allowing long sequences to be processed with substantially better memory
scalability. Reducing peak GPU memory does not make \smr{} computationally free;
instead, it trades a bounded GPU working set for additional memory
management. Table~\ref{tab:compute} shows that computation in GPU
remain dominated by the underlying geometry foundation model, while
\smr{} adds modest CPU, PFLOP, and runtime overhead. In return, \smr{}
enables substantially better memory scaling; for example, STream3R runs
out of GPU memory at $N=100$, whereas STream3R+$\smr{}$ can still process
the full sequence using bounded-window inference.

\section{Conclusion and Future Work}
\label{sec:discussion}
We introduced \smr{}, a revisable external memory that extends frozen
geometry foundation models beyond fixed-context prediction through
storage, retrieval, geometric verification, and revision. Across eight
geometry models, five datasets, and five downstream tasks, our results
show that \smr{} complements both feed-forward and streaming models,
using verified revisits to correct accumulated errors over long visual
sequences without fine-tuning the underlying geometry models.
Several directions remain for future work. Retrieval verification
introduces additional computation, motivating more efficient mechanisms
for recalling reliable spatial information from the memory scaffold.
Moreover, \smr{} currently focuses on revising global window placement;
future work could extend revision to local geometric predictions when
later observations provide stronger evidence.
\label{page:main-end}

\FloatBarrier
\bibliographystyle{iclr2027_conference}
\bibliography{references}

@article{shen2026learning,
  title={Learning to Perceive" Where": Spatial Pretext Tasks for Robust Self-Supervised Learning},
  author={Shen, Yang and Cai, Yusen and Hryniewska-Guzik, Weronika and Lin, Qing and Zhang, Mengmi},
  journal={arXiv preprint arXiv:2605.09963},
  year={2026}
}

@article{khandelwal2023adaptive,
  title={Adaptive visual scene understanding: incremental scene graph generation},
  author={Khandelwal, Naitik and Liu, Xiao and Zhang, Mengmi},
  journal={arXiv preprint arXiv:2310.01636},
  year={2023}
}

@inproceedings{zhang2026peering,
  title={Peering into the unknown: Active view selection with neural uncertainty maps for 3d reconstruction},
  author={Zhang, Zhengquan and Xu, Feng and Zhang, Mengmi},
  booktitle={International Conference on Learning Representations},
  volume={2026},
  pages={85781--85801},
  year={2026}
}

@article{piriyajitakonkij2024seeing,
  title={Seeing Through Uncertainty: A Free-Energy Approach for Real-Time Perceptual Adaptation in Robust Visual Navigation},
  author={Piriyajitakonkij, Maytus and Yadav, Rishabh Dev and Sun, Mingfei and Zhang, Mengmi and Pan, Wei},
  journal={arXiv preprint arXiv:2403.01977},
  year={2024}
}

@inproceedings{zhangegocentric,
  title={Egocentric spatial memory},
  author={Zhang, Mengmi and Ma, Keng Teck and Yen, Shih-Cheng and Lim, Joo Hwee and Zhao, Qi and Feng, Jiashi},
  booktitle={2018 IEEE/RSJ International Conference on Intelligent Robots and Systems (IROS)},
  pages={137--144},
  year={2018},
  organization={IEEE}
}

@article{zhang2018anticipating,
  title={Anticipating where people will look using adversarial networks},
  author={Zhang, Mengmi and Ma, Keng Teck and Lim, Joo Hwee and Zhao, Qi and Feng, Jiashi},
  journal={IEEE transactions on pattern analysis and machine intelligence},
  volume={41},
  number={8},
  pages={1783--1796},
  year={2018},
  publisher={IEEE}
}

@inproceedings{vggt,
  title={{VGGT}: Visual Geometry Grounded Transformer},
  author={Wang, Jianyuan and Chen, Minghao and Karaev, Nikita and Vedaldi, Andrea and Rupprecht, Christian and Novotny, David},
  booktitle={Proceedings of the IEEE/CVF Conference on Computer Vision and Pattern Recognition (CVPR)},
  pages={5294--5306},
  year={2025}
}

@inproceedings{vggtomega,
  title={{VGGT-$\Omega$}},
  author={Wang, Jianyuan and Chen, Minghao and Zhang, Shangzhan and Karaev, Nikita and Sch{\"o}nberger, Johannes and Labatut, Patrick and Bojanowski, Piotr and Novotny, David and Vedaldi, Andrea and Rupprecht, Christian},
  booktitle={Proceedings of the IEEE/CVF Conference on Computer Vision and Pattern Recognition (CVPR)},
  year={2026}
}

@inproceedings{pi3,
  title={$\pi^3$: Permutation-Equivariant Visual Geometry Learning},
  author={Wang, Yifan and Zhou, Jianjun and Zhu, Haoyi and Chang, Wenzheng and Zhou, Yang and Li, Zizun and Chen, Junyi and Pang, Jiangmiao and Shen, Chunhua and He, Tong},
  booktitle={International Conference on Learning Representations (ICLR)},
  year={2026}
}

@inproceedings{dust3r,
  title={{DUSt3R}: Geometric 3D Vision Made Easy},
  author={Wang, Shuzhe and Leroy, Vincent and Cabon, Yohann and Chidlovskii, Boris and Revaud, J{\'e}r{\^o}me},
  booktitle={Proceedings of the IEEE/CVF Conference on Computer Vision and Pattern Recognition (CVPR)},
  pages={20697--20709},
  year={2024}
}

@inproceedings{mast3r,
  title={Grounding Image Matching in 3D with {MASt3R}},
  author={Leroy, Vincent and Cabon, Yohann and Revaud, J{\'e}r{\^o}me},
  booktitle={European Conference on Computer Vision (ECCV)},
  pages={71--91},
  year={2024}
}

@inproceedings{fast3r,
  title={{Fast3R}: Towards 3D Reconstruction of 1000+ Images in One Forward Pass},
  author={Yang, Jianing and Sax, Alexander and Liang, Kevin J. and Henaff, Mikael and Tang, Hao and Cao, Ang and Chai, Joyce and Meier, Franziska and Feiszli, Matt},
  booktitle={Proceedings of the IEEE/CVF Conference on Computer Vision and Pattern Recognition (CVPR)},
  pages={21924--21935},
  year={2025}
}

@article{vectorhash,
  title={Episodic and associative memory from spatial scaffolds in the hippocampus},
  author={Chandra, Sarthak and Sharma, Sugandha and Chaudhuri, Rishidev and Fiete, Ila},
  journal={Nature},
  volume={638},
  pages={739--751},
  year={2025},
  doi={10.1038/s41586-024-08392-y}
}

@article{dinov2,
  title={{DINOv2}: Learning Robust Visual Features without Supervision},
  author={Oquab, Maxime and Darcet, Timoth{\'e}e and Moutakanni, Th{\'e}o and Vo, Huy V. and Szafraniec, Marc and Khalidov, Vasil and Fernandez, Pierre and Haziza, Daniel and Massa, Francisco and El-Nouby, Alaaeldin and Assran, Mido and Ballas, Nicolas and Galuba, Wojciech and Howes, Russell and Huang, Po-Yao and Li, Shang-Wen and Misra, Ishan and Rabbat, Michael and Sharma, Vasu and Synnaeve, Gabriel and Xu, Hu and J{\'e}gou, Herv{\'e} and Mairal, Julien and Labatut, Patrick and Joulin, Armand and Bojanowski, Piotr},
  journal={Transactions on Machine Learning Research},
  year={2024}
}

@article{umeyama,
  title={Least-squares estimation of transformation parameters between two point patterns},
  author={Umeyama, Shinji},
  journal={IEEE Transactions on Pattern Analysis and Machine Intelligence},
  volume={13},
  number={4},
  pages={376--380},
  year={1991},
  doi={10.1109/34.88573}
}

@article{horn,
  title={Closed-form solution of absolute orientation using unit quaternions},
  author={Horn, Berthold K. P.},
  journal={Journal of the Optical Society of America A},
  volume={4},
  number={4},
  pages={629--642},
  year={1987},
  doi={10.1364/JOSAA.4.000629}
}

@inproceedings{strasdat,
  title={Scale Drift-Aware Large Scale Monocular {SLAM}},
  author={Strasdat, Hauke and Montiel, J. M. M. and Davison, Andrew J.},
  booktitle={Robotics: Science and Systems (RSS)},
  year={2010},
  doi={10.15607/RSS.2010.VI.010}
}

@article{grisetti,
  title={A Tutorial on Graph-Based {SLAM}},
  author={Grisetti, Giorgio and K{\"u}mmerle, Rainer and Stachniss, Cyrill and Burgard, Wolfram},
  journal={IEEE Intelligent Transportation Systems Magazine},
  volume={2},
  number={4},
  pages={31--43},
  year={2010},
  doi={10.1109/MITS.2010.939925}
}

@article{huber,
  title={Robust Estimation of a Location Parameter},
  author={Huber, Peter J.},
  journal={The Annals of Mathematical Statistics},
  volume={35},
  number={1},
  pages={73--101},
  year={1964},
  doi={10.1214/aoms/1177703732}
}

@inproceedings{elasticfusion,
  title={{ElasticFusion}: Dense {SLAM} without a Pose Graph},
  author={Whelan, Thomas and Leutenegger, Stefan and Salas-Moreno, Renato F. and Glocker, Ben and Davison, Andrew J.},
  booktitle={Robotics: Science and Systems (RSS)},
  year={2015},
  doi={10.15607/RSS.2015.XI.001}
}

@inproceedings{sevenscenes,
  title={Scene Coordinate Regression Forests for Camera Relocalization in {RGB-D} Images},
  author={Shotton, Jamie and Glocker, Ben and Zach, Christopher and Izadi, Shahram and Criminisi, Antonio and Fitzgibbon, Andrew},
  booktitle={Proceedings of the IEEE Conference on Computer Vision and Pattern Recognition (CVPR)},
  pages={2930--2937},
  year={2013}
}

@inproceedings{streamvggt,
  title={Streaming Visual Geometry Transformer},
  author={Zhuo, Dong and Zheng, Wenzhao and Guo, Jiahe and Wu, Yuqi and Zhou, Jie and Lu, Jiwen},
  booktitle={International Conference on Learning Representations (ICLR)},
  year={2026}
}

@inproceedings{stream3r,
  title={{STream3R}: Scalable Sequential {3D} Reconstruction with Causal Transformer},
  author={Lan, Yushi and Luo, Yihang and Hong, Fangzhou and Zhou, Shangchen and Chen, Honghua and Lyu, Zhaoyang and Yang, Shuai and Dai, Bo and Loy, Chen Change and Pan, Xingang},
  booktitle={International Conference on Learning Representations (ICLR)},
  year={2026}
}

@inproceedings{cut3r,
  title={Continuous 3D Perception Model with Persistent State},
  author={Wang, Qianqian and Zhang, Yifei and Holynski, Aleksander and Efros, Alexei A. and Kanazawa, Angjoo},
  booktitle={Proceedings of the IEEE/CVF Conference on Computer Vision and Pattern Recognition (CVPR)},
  pages={10510--10522},
  year={2025}
}

@inproceedings{nicerslam,
  title={{NICER-SLAM}: Neural Implicit Scene Encoding for {RGB} {SLAM}},
  author={Zhu, Zihan and Peng, Songyou and Larsson, Viktor and Cui, Zhaopeng and Oswald, Martin R. and Geiger, Andreas and Pollefeys, Marc},
  booktitle={International Conference on 3D Vision (3DV)},
  pages={42--52},
  year={2024}
}

@inproceedings{droidslam,
  title={{DROID-SLAM}: Deep Visual {SLAM} for Monocular, Stereo, and {RGB-D} Cameras},
  author={Teed, Zachary and Deng, Jia},
  booktitle={Advances in Neural Information Processing Systems (NeurIPS)},
  volume={34},
  year={2021}
}

@inproceedings{mast3rslam,
  title={{MASt3R-SLAM}: Real-Time Dense {SLAM} with {3D} Reconstruction Priors},
  author={Murai, Riku and Dexheimer, Eric and Davison, Andrew J.},
  booktitle={Proceedings of the IEEE/CVF Conference on Computer Vision and Pattern Recognition (CVPR)},
  pages={16695--16705},
  year={2025}
}

@inproceedings{vggtslam,
  title={{VGGT-SLAM}: Dense {RGB} {SLAM} Optimized on the {SL(4)} Manifold},
  author={Maggio, Dominic and Lim, Hyungtae and Carlone, Luca},
  booktitle={Advances in Neural Information Processing Systems (NeurIPS)},
  volume={38},
  year={2025}
}

@inproceedings{vggtslampp,
  title={{VGGT-SLAM++}},
  author={Mandal, Avilasha and Kumar, Rajesh and Harithas, Sudarshan Sunil and Arora, Chetan},
  booktitle={Proceedings of the IEEE/CVF Conference on Computer Vision and Pattern Recognition (CVPR) Workshops},
  pages={1008--1018},
  year={2026}
}

@inproceedings{vistaslam,
  title={{ViSTA-SLAM}: Visual {SLAM} with Symmetric Two-view Association},
  author={Zhang, Ganlin and Qian, Shenhan and Wang, Xi and Cremers, Daniel},
  booktitle={International Conference on 3D Vision (3DV)},
  pages={396--406},
  year={2026}
}

@inproceedings{slamformer,
  title={{SLAM-Former}: Putting {SLAM} into One Transformer},
  author={Yuan, Yijun and Chen, Zhuoguang and Li, Kenan and Wang, Weibang and Qin, Minghui and Fang, Zhijian and Zheng, Weicheng and Zhao, Hang},
  booktitle={European Conference on Computer Vision (ECCV)},
  year={2026}
}

@article{kerbl3dgs,
  title={{3D} Gaussian Splatting for Real-Time Radiance Field Rendering},
  author={Kerbl, Bernhard and Kopanas, Georgios and Leimk{\"u}hler, Thomas and Drettakis, George},
  journal={ACM Transactions on Graphics},
  volume={42},
  number={4},
  year={2023},
  doi={10.1145/3592433}
}

@inproceedings{lpips,
  title={The Unreasonable Effectiveness of Deep Features as a Perceptual Metric},
  author={Zhang, Richard and Isola, Phillip and Efros, Alexei A. and Shechtman, Eli and Wang, Oliver},
  booktitle={Proceedings of the IEEE Conference on Computer Vision and Pattern Recognition (CVPR)},
  pages={586--595},
  year={2018}
}

@inproceedings{co3d,
  title={Common objects in 3d: Large-scale learning and evaluation of real-life 3d category reconstruction},
  author={Reizenstein, Jeremy and Shapovalov, Roman and Henzler, Philipp and Sbordone, Luca and Labatut, Patrick and Novotny, David},
  booktitle={2021 IEEE/CVF International Conference on Computer Vision (ICCV)},
  pages={10881--10891},
  year={2021},
  organization={IEEE}
}

@article{re10k,
  title={Stereo magnification: Learning view synthesis using multiplane images},
  author={Zhou, Tinghui and Tucker, Richard and Flynn, John and Fyffe, Graham and Snavely, Noah},
  journal={arXiv preprint arXiv:1805.09817},
  year={2018}
}

@inproceedings{talo,
  title={Talo: Pushing 3d vision foundation models towards globally consistent online reconstruction},
  author={Zhang, Fengyi and Zhang, Tianjun and Khosoussi, Kasra and Zhang, Zheng and Huang, Zi and Luo, Yadan},
  booktitle={Proceedings of the IEEE/CVF Conference on Computer Vision and Pattern Recognition},
  pages={21870--21879},
  year={2026}
}

@inproceedings{amb3r,
  title={Amb3r: Accurate feed-forward metric-scale 3d reconstruction with backend},
  author={Wang, Hengyi and Agapito, Lourdes},
  booktitle={Proceedings of the IEEE/CVF Conference on Computer Vision and Pattern Recognition},
  pages={14612--14625},
  year={2026}
}

@article{gluemap,
  title={Global Structure-from-Motion Meets Feedforward Reconstruction},
  author={Pan, Linfei and Sch{\"o}nberger, Johannes and Pollefeys, Marc},
  journal={arXiv preprint arXiv:2605.26103},
  year={2026}
}

@article{eagle,
  title={EAGLE: Episodic Appearance-and Geometry-aware Memory for Unified 2D-3D Visual Query Localization in Egocentric Vision},
  author={Cao, Yifei and Liu, Yu and Wang, Guolong and Liu, Zhu and Wang, Kai and Zhang, Xianjie and Yu, Jizhe and Tu, Xun},
  journal={Proceedings of the AAAI Conference on Artificial Intelligence},
  volume={40},
  number={4},
  pages={2634--2642},
  year={2026}
}

@inproceedings{vggsfm,
  title={Vggsfm: Visual geometry grounded deep structure from motion},
  author={Wang, Jianyuan and Karaev, Nikita and Rupprecht, Christian and Novotny, David},
  booktitle={2024 IEEE/CVF Conference on Computer Vision and Pattern Recognition (CVPR)},
  pages={21686--21697},
  year={2024},
  organization={IEEE}
}

@article{gsplat,
  title={gsplat: An open-source library for Gaussian splatting},
  author={Ye, Vickie and Li, Ruilong and Kerr, Justin and Turkulainen, Matias and Yi, Brent and Pan, Zhuoyang and Seiskari, Otto and Ye, Jianbo and Hu, Jeffrey and Tancik, Matthew and others},
  journal={Journal of Machine Learning Research},
  volume={26},
  number={34},
  pages={1--17},
  year={2025}
}

@article{amari1977dynamics,
  title={Dynamics of pattern formation in lateral-inhibition type neural fields},
  author={Amari, Shun-ichi},
  journal={Biological cybernetics},
  volume={27},
  number={2},
  pages={77--87},
  year={1977},
  publisher={Springer}
}

@article{burak2009accurate,
  title={Accurate path integration in continuous attractor network models of grid cells},
  author={Burak, Yoram and Fiete, Ila R},
  journal={PLoS computational biology},
  volume={5},
  number={2},
  pages={e1000291},
  year={2009},
  publisher={Public Library of Science San Francisco, USA}
}

@article{bicanski2020neuronal,
  title={Neuronal vector coding in spatial cognition},
  author={Bicanski, Andrej and Burgess, Neil},
  journal={Nature Reviews Neuroscience},
  volume={21},
  number={9},
  pages={453--470},
  year={2020},
  publisher={Nature Publishing Group UK London}
}

@inproceedings{mixvpr,
  title={Mixvpr: Feature mixing for visual place recognition},
  author={Ali-Bey, Amar and Chaib-Draa, Brahim and Giguere, Philippe},
  booktitle={2023 IEEE/CVF Winter Conference on Applications of Computer Vision (WACV)},
  pages={2997--3006},
  year={2023},
  organization={IEEE}
}

@inproceedings{netvlad,
  title={NetVLAD: CNN architecture for weakly supervised place recognition},
  author={Arandjelovic, Relja and Gronat, Petr and Torii, Akihiko and Pajdla, Tomas and Sivic, Josef},
  booktitle={Proceedings of the IEEE conference on computer vision and pattern recognition},
  pages={5297--5307},
  year={2016}
}

@inproceedings{eigenplaces,
  title={Eigenplaces: Training viewpoint robust models for visual place recognition},
  author={Berton, Gabriele and Trivigno, Gabriele and Caputo, Barbara and Masone, Carlo},
  booktitle={2023 IEEE/CVF International Conference on Computer Vision (ICCV)},
  pages={11046--11056},
  year={2023},
  organization={IEEE}
}

@inproceedings{kummerle2011g,
  title={g 2 o: A general framework for graph optimization},
  author={K{\"u}mmerle, Rainer and Grisetti, Giorgio and Strasdat, Hauke and Konolige, Kurt and Burgard, Wolfram},
  booktitle={2011 IEEE international conference on robotics and automation},
  pages={3607--3613},
  year={2011},
  organization={IEEE}
}

@techreport{dellaert2012factor,
  title={Factor graphs and GTSAM: A hands-on introduction},
  author={Dellaert, Frank},
  year={2012},
  number={GT-RIM-CP\&R-2012-002},
  institution={Georgia Institute of Technology}
}

@inproceedings{cosplace,
  title={Rethinking visual geo-localization for large-scale applications},
  author={Berton, Gabriele and Masone, Carlo and Caputo, Barbara},
  booktitle={2022 IEEE/CVF Conference on Computer Vision and Pattern Recognition (CVPR)},
  pages={4868--4878},
  year={2022},
  organization={IEEE}
}

@inproceedings{salad,
  title={Optimal transport aggregation for visual place recognition},
  author={Izquierdo, Sergio and Civera, Javier},
  booktitle={2024 IEEE/CVF Conference on Computer Vision and Pattern Recognition (CVPR)},
  pages={17658--17668},
  year={2024},
  organization={IEEE}
}

@inproceedings{boq,
  title={BoQ: A place is worth a bag of learnable queries},
  author={Ali-Bey, Amar and Chaib-Draa, Brahim and Giguere, Philippe},
  booktitle={2024 IEEE/CVF Conference on Computer Vision and Pattern Recognition (CVPR)},
  pages={17794--17803},
  year={2024},
  organization={IEEE}
}

@article{vggtslam2,
  title={VGGT-SLAM 2.0: Real-time dense feed-forward scene reconstruction},
  author={Maggio, Dominic and Carlone, Luca},
  journal={arXiv preprint arXiv:2601.19887},
  year={2026}
}

@article{hu2022real,
  title={A real-time map restoration algorithm based on ORB-SLAM3},
  author={Hu, Weiwei and Lin, Qinglei and Shao, Lihuan and Lin, Jiaxu and Zhang, Keke and Qin, Huibin},
  journal={Applied Sciences},
  volume={12},
  number={15},
  pages={7780},
  year={2022},
  publisher={MDPI}
}

@article{replica,
  title={The replica dataset: A digital replica of indoor spaces},
  author={Straub, Julian and Whelan, Thomas and Ma, Lingni and Chen, Yufan and Wijmans, Erik and Green, Simon and Engel, Jakob J and Mur-Artal, Raul and Ren, Carl and Verma, Shobhit and others},
  journal={arXiv preprint arXiv:1906.05797},
  year={2019}
}

@inproceedings{scannetpp,
  title={Scannet++: A high-fidelity dataset of 3d indoor scenes},
  author={Yeshwanth, Chandan and Liu, Yueh-Cheng and Nie{\ss}ner, Matthias and Dai, Angela},
  booktitle={2023 IEEE/CVF International Conference on Computer Vision (ICCV)},
  pages={12--22},
  year={2023},
  organization={IEEE}
}
\appendix

\setcounter{figure}{0}
\renewcommand{\thefigure}{S\arabic{figure}}

\setcounter{table}{0}
\renewcommand{\thetable}{S\arabic{table}}

\renewcommand{\theHfigure}{S\arabic{figure}}
\renewcommand{\theHtable}{S\arabic{table}}

\clearpage
\section{Appendix Roadmap}
\label{app:overview}

The appendix provides the mathematical details, evaluation protocols,
and diagnostics supporting the main paper. Appendices~\ref{app:sim3}--\ref{app:pgo} describe the memory and geometry; Appendices~\ref{app:datasets}--\ref{app:nvs}
give the evaluation and rendering protocols; the remaining sections report
ablations, resource costs, implementation settings, and failure analyses.
The links below lead directly to each section.

\begingroup
\small
\renewcommand{\arraystretch}{1.3}
\begin{tabularx}{\linewidth}{@{}lXr@{}}
\toprule
\textbf{Appendix} & \textbf{Contents} & \textbf{Page}\\
\midrule
\ref{app:sim3} & \hyperref[app:sim3]{Camera-pose alignment} & \pageref{app:sim3}\\
\ref{app:invariant} & \hyperref[app:invariant]{Window ownership and geometric invariance} & \pageref{app:invariant}\\
\ref{app:vhash} & \hyperref[app:vhash]{Scaffold construction, storage, and retrieval} & \pageref{app:vhash}\\
\ref{app:association} & \hyperref[app:association]{Scaffold dynamics and controlled memory diagnostics} & \pageref{app:association}\\
\ref{app:closure} & \hyperref[app:closure]{Geometric verification of retrieved revisits} & \pageref{app:closure}\\
\ref{app:pgo} & \hyperref[app:pgo]{Trajectory revision and global refinement} & \pageref{app:pgo}\\
\ref{app:datasets} & \hyperref[app:datasets]{Datasets and evaluation protocols} & \pageref{app:datasets}\\
\ref{app:nvs} & \hyperref[app:nvs]{Novel view synthesis protocol and decoder} & \pageref{app:nvs}\\
\ref{app:ablations} & \hyperref[app:ablations]{Controlled ablations} & \pageref{app:ablations}\\
\ref{app:compute} & \hyperref[app:compute]{Compute and memory measurements} & \pageref{app:compute}\\
\ref{app:repro} & \hyperref[app:repro]{Implementation settings and reproducibility} & \pageref{app:repro}\\
\ref{app:anchored-distortion} & \hyperref[app:anchored-distortion]{Local-prediction changes and trajectory revision} & \pageref{app:anchored-distortion}\\
\ref{app:bank-growth} & \hyperref[app:bank-growth]{Memory growth and pruning} & \pageref{app:bank-growth}\\
\bottomrule
\end{tabularx}
\endgroup

\section{Camera-Pose Alignment}
\label{app:sim3}

The main text uses camera-pose alignment in two places. First, in the opening of
Sec.~\ref{sec:method}, consecutive windows are aligned using
the camera poses of their shared frames to obtain their global placements
$T_k$. Second, in Sec.~\ref{sec:retrieval-check}, the newly predicted
poses of a retrieved pair are aligned with the stored poses of the same
frames in their source window to obtain the relative transformation
$\widehat Z_{o,k}$. Both operations use the same similarity-alignment
procedure described below.

\paragraph{Similarity transformation.}
Suppose the same frames have camera predictions in two coordinate
systems, $A$ and $B$. We seek the transformation that expresses the
predictions from $A$ in coordinate system $B$. We write this similarity
transformation as
\[
S=(s,R,t),
\]
where $s>0$ is scale, $R\in\mathrm{SO}(3)$ is rotation, and
$t\in\mathbb R^3$ is translation. For a 3D point $x$,
\[
S\cdot x=sRx+t.
\]
For a camera pose $(R_j,c_j)$, where $R_j$ is its orientation and $c_j$
its camera centre,
\[
S\cdot(R_j,c_j)
=
\left(
RR_j,\;
sRc_j+t
\right).
\]

\paragraph{Estimating the alignment.}
Because the two predictions correspond to the same frames, their camera
centres provide known correspondences
$c_j^A\leftrightarrow c_j^B$. We estimate the transformation from
coordinate system $A$ to $B$ by
\begin{equation}
\widehat S_{A\rightarrow B}
=
\argmin_{s>0,\,R\in\mathrm{SO}(3),\,t}
\sum_j
w_j
\left\|
sRc_j^A+t-c_j^B
\right\|_2^2 .
\label{eq:sim3-alignment}
\end{equation}
Thus, no additional frame matching or nearest-neighbour search is needed.
Scale, rotation, and translation are estimated jointly from the
corresponding camera centres using standard similarity
alignment~\citep{umeyama,horn}.

This is the operation used in Sec.~\ref{sec:method}: for two
consecutive windows, the corresponding cameras are those of their shared
frames. The resulting relative transformation places the new window with
respect to the preceding window, and composing these transformations gives
its global placement $T_k$.

The same operation is used in Sec.~\ref{sec:retrieval-check}. For a
retrieved pair from source window $\mathcal W_o$, the two cameras predicted
in the joint pass are aligned with the stored poses of the same two frames
in $\mathcal W_o$. This gives
\[
\widehat Z_{o,k}
=
\widehat S_{\mathrm{joint}\rightarrow o},
\]
which relates the current joint prediction to the source window. The
source window already has global placement $T_o$, so the corresponding
proposal for the current window is
\[
T_k^{(a)}=T_o\widehat Z_{o,k}.
\]

\paragraph{Making the alignment robust.}
Camera predictions can contain outliers, so we refine the similarity fit
for five reweighting rounds. After each round, the centre residual is
\[
r_j=
\left\|
\widehat S_{A\rightarrow B}\cdot c_j^A-c_j^B
\right\|_2 ,
\]
and the weight used in the next round is
\[
w_j
=
\min\!\left(
1,\;
\kappa\frac{\widetilde r}{r_j}
\right),
\qquad
\kappa=2.5,
\]
where $\widetilde r$ is the median residual. A zero-residual
correspondence keeps unit weight. After fitting the camera centres, we
also compare the transformed camera orientation $RR_j^A$ with $R_j^B$
and reject correspondences with excessive rotational disagreement.

\paragraph{Additional scale check for consecutive windows.}
For consecutive-window alignment, an incorrect scale would affect every
later window in the trajectory. We therefore compare the fitted scale
$s$ with an estimate $s_{\mathrm{pair}}$ obtained from the median ratio
of pairwise camera-centre distances. We accept the alignment only if
\[
\frac{s}{s_{\mathrm{pair}}}
\in
\left[
\frac{1}{\gamma_s},
\gamma_s
\right],
\qquad
\gamma_s=1.5.
\]
This scale check is applied to the sequential window alignment described
in Sec.~\ref{sec:method}.

\paragraph{Two-camera limitation.}
The shared-frame alignment between consecutive windows normally has
multiple camera correspondences. A retrieved pair in
Sec.~\ref{sec:retrieval-check}, however, contains only two cameras.
Two camera centres determine a baseline but do not by themselves uniquely
constrain rotation about that baseline. The camera-orientation check
provides additional consistency information, and \smr{} further requires
two retrieved pairs from different source windows to produce consistent
global-placement proposals before accepting a revisit.

\section{Window Ownership and Geometric Invariance}
\label{app:invariant}

In Sec.~\ref{sec:memory-storage}, each frame is stored once using its
source window, and its camera pose and point map remain in that window's
local coordinate system. Later, Sec.~\ref{sec:placement-update} changes
only the global placement $T_k$ of a window. Here we make explicit what
this means for the stored geometry and for relative camera poses.

\paragraph{Source-window storage.}
Because consecutive windows overlap, the same frame may be predicted in
more than one window. We store it only once, using the first window in
which it appears. For frame $I_i$, this window is
$\mathcal W_{\operatorname{src}(i)}$.

Suppose $I_i$ is stored in source window $\mathcal W_k$, with local
camera pose $(R_i,c_i)$ and local 3D points $x_i$. These quantities remain
fixed in the coordinate system of $\mathcal W_k$. Their global values are
obtained only when the window placement
$T_k=(a_k,Q_k,\mathbf b_k)$ is applied:
\[
R_i^{\mathrm{global}} = Q_kR_i,
\qquad
c_i^{\mathrm{global}}
=
a_kQ_kc_i+\mathbf b_k,
\]
and
\[
x_i^{\mathrm{global}}
=
a_kQ_kx_i+\mathbf b_k.
\]

Thus, the content bank stores local geometry, while $T_k$ determines
where that geometry currently lies in the global reconstruction.

\paragraph{What changes after a placement correction?}
When Sec.~\ref{sec:placement-update} replaces $T_k$ by an updated global
placement, the stored local camera poses and point maps are not
recomputed. Instead, the new transformation is applied to the same local
quantities. Consequently, all cameras and 3D points belonging to
$\mathcal W_k$ move together.

This separation is useful because a revisit can correct where a window
is placed without modifying the geometric relationships already predicted
inside that window.

\paragraph{Relative poses within one source window.}
Consider two frames $I_i$ and $I_j$ stored in the same source window
$\mathcal W_k$. After applying $T_k$, their global rotations satisfy
\[
\left(R_j^{\mathrm{global}}\right)^\top
R_i^{\mathrm{global}}
=
R_j^\top R_i.
\]
Their relative translation, expressed in frame $I_j$, becomes
\[
\left(R_j^{\mathrm{global}}\right)^\top
\left(
c_i^{\mathrm{global}}-c_j^{\mathrm{global}}
\right)
=
a_k R_j^\top(c_i-c_j).
\]
Therefore, changing $T_k$ leaves their relative rotation and translation
direction unchanged. Only the translation magnitude can change through
the window scale $a_k$.

This is why a placement correction primarily changes relationships
\emph{between} windows rather than the relative camera geometry already
stored \emph{within} one source window. It also explains why the
pairwise-pose metric, which uses rotation and translation-direction
errors, is unchanged for two frames whose local predictions remain fixed
in the same source window.

\paragraph{When this property does not apply.}
The argument above assumes that the stored local predictions
$(R_i,c_i)$ themselves are unchanged. In
Sec.~\ref{sec:retrieval-check}, however, processing retrieved frames
jointly with the current window can also produce new local camera poses
and point maps. That is a separate effect from changing $T_k$.
Appendix~\ref{app:anchored-distortion} isolates these two sources of
change experimentally.

\section{Scaffold Construction, Storage, and Retrieval}
\label{app:vhash}

Sections~\ref{sec:memory-storage} and~\ref{sec:memory-retrieval}
introduce the two memory operations used by \smr{}. When a frame is
stored, its estimated global camera pose is converted into a compact
memory address. Later, the visual content of a current frame is used to
predict an address and retrieve previously stored frames. Here we explain
how the pose representation, memory address, and visual descriptor are
connected, why the scaffold uses multiple spatial scales and a sparse
address, and which parts of this associative pathway are actually used
during trajectory revision.

\paragraph{Representing the global camera pose.}
When frame $I_i$ is first stored, its source window has a current global
placement $T_{\operatorname{src}(i)}$. Applying this placement to the
frame's local camera pose gives the global camera pose used to construct
its scaffold representation. As described in
Sec.~\ref{sec:memory-storage}, we represent camera position and
orientation separately and concatenate them into a single scaffold state
$\mathbf g_i$.

Camera position is first expressed in normalized scene units, with
median scene depth equal to one. We use three spatial modules with
periods
\[
\lambda_1=2.4,\qquad
\lambda_2=3.2,\qquad
\lambda_3=4.0.
\]
For module $m$, a position
$\mathbf p_i=(x_i,y_i,z_i)$ is converted to periodic coordinates
\[
\boldsymbol{\phi}_{m,i}
=
\frac{2\pi}{\lambda_m}
(x_i,y_i,z_i)
\bmod 2\pi .
\]
Thus, each individual module is periodic: positions separated by
$\lambda_m$ along an encoded axis produce the same phase in that module.

A single periodic module cannot distinguish locations separated by its
period. The three modules reduce this ambiguity because their periods are
different: a displacement that wraps one module generally does not wrap
the other two at the same location. We use
\[
(2.4,3.2,4.0)=0.8(3,4,5),
\]
so the complete three-module pattern repeats only after
\[
0.8\,\operatorname{lcm}(3,4,5)=48
\]
normalized units along an encoded axis. The representation is therefore
less ambiguous than any individual module, although it is still
periodic rather than globally unique over an unbounded scene. These
periods are fixed across the main experiments; we do not claim that
they are an optimal set of spatial periods.

For each spatial module, the horizontal phase is represented on a
$48\times48$ periodic grid and height on a 48-unit ring. These numbers
specify the resolution with which a module represents its phase; they
are distinct from the spatial period $\lambda_m$. We use
$48\times48$ as the default horizontal resolution. The sensitivity study
in Table~\ref{tab:ablation-index} additionally evaluates $16\times16$ and
$64\times64$ modules and shows that the method does not depend on using
the finest tested resolution.

\paragraph{Representing camera orientation.}
Camera orientation is converted to yaw, pitch, and roll using the ZYX
convention. Each angle is represented on a separate 256-unit periodic
ring. A ring is natural for an angular variable because the beginning
and end of the interval correspond to the same orientation after a
full $360^\circ$ rotation.

The three spatial modules and three orientation rings are concatenated,
giving
\[
\mathbf g_i\in\mathbb R^{N_g},
\qquad
N_g
=
3(48^2+48)+3(256)
=
7{,}824.
\]
Thus, $\mathbf g_i$ is the scaffold representation of \emph{where}
frame $I_i$ was estimated to have been captured when it entered memory.

\paragraph{From the scaffold state to a compact memory address.}
Directly associating visual descriptors with the full
$7{,}824$-dimensional scaffold state would make the associative layer
unnecessarily large. We therefore project the scaffold state into a
smaller address space:
\[
\mathbf z_i=W_{gh}\mathbf g_i,
\qquad
W_{gh}\in\mathbb R^{1024\times7824}.
\]
The entries of $W_{gh}$ are sampled independently from
$\mathcal N(0,1/7824)$ and the matrix remains fixed throughout
inference.

We then retain only the largest 64 entries:
\[
\mathbf v_i
=
\sigma_{64}(\mathbf z_i),
\qquad
\mathbf h_i
=
\frac{\mathbf v_i}{\|\mathbf v_i\|_2},
\]
where $\sigma_{64}$ sets all but the 64 largest values to zero. The
resulting
\[
\mathbf h_i\in\mathbb R^{1024}
\]
is the memory address of frame $I_i$.

The purpose of the top-$k$ operation is to make different addresses
selective: unrelated scaffold states activate only a small subset of
the 1024 address units, reducing overlap between stored
representations. The main configuration retains $k=64$ units, i.e.,
$6.25\%$ of the address layer. We use this as a fixed sparsity setting
throughout the main experiments. The address dimension is
examined in Table~\ref{tab:ablation-index}: increasing it from the default
$N_h=1024$ to $8192$ does not consistently improve the final results,
while reducing it to $512$ produces only a modest change. Thus, the
reported gains do not rely on an exceptionally large address layer.

Normalizing $\mathbf h_i$ also makes comparisons between addresses
depend on their represented direction and overlap rather than on the
absolute magnitude produced by the random projection.

The memory address does not contain the RGB frame or its dense geometry.
Instead, $\mathbf h_i$ is linked to the corresponding content-bank entry.
That entry stores the RGB frame, its source window, and its camera pose
and point map in the local coordinate system of that source window.
Appendix~\ref{app:invariant} explains how the current global placement
$T_{\operatorname{src}(i)}$ is applied when this content is used. The address is assigned when the frame is first stored. Later placement
corrections update $T_{\operatorname{src}(i)}$ but do not recompute
$\mathbf h_i$. This separation allows the stored local geometry to move
with its source window without rewriting the content bank or its memory
index.

\paragraph{Recovering a scaffold state from an address.}
The mapping above provides the direction
$\mathbf g\rightarrow\mathbf h$. The scaffold additionally contains a
return mapping
\[
W_{hg}\in\mathbb R^{7824\times1024}
\]
for the reverse direction. Given an address,
\[
\widehat{\mathbf g}_i
=
\psi(W_{hg}\mathbf h_i),
\]
where $\psi$ normalizes each spatial or orientation field independently.

Unlike $W_{gh}$, which is a fixed random projection, $W_{hg}$ is fitted
once before inference. We sample valid scaffold states
$\{\mathbf g_p\}$, construct their corresponding addresses
$\{\mathbf h_p\}$ using the same $W_{gh}$ and sparsification rule, and
solve
\[
W_{hg}
=
\argmin_W
\sum_p
\left\|
\mathbf g_p-W\mathbf h_p
\right\|_2^2
+
\lambda_{\mathrm{ridge}}\|W\|_F^2.
\]
The sampled scaffold states use a maximum phase spacing of
$\lambda_{\min}/6=0.4$. This fitting procedure uses no RGB frames,
geometry-foundation-model outputs, or evaluation labels. It prepares the
scaffold's return mapping once and is separate from the online memory
updates described below.

The spatial part of the associative memory can therefore be written as
\[
\mathbf g_i
\;\xrightleftharpoons[W_{hg}]{W_{gh}}\;
\mathbf h_i .
\]

\paragraph{Associating visual content with the same address.}
The scaffold tells us \emph{where} a stored frame was captured, but at
test time we begin with \emph{what} the current frame looks like.
Therefore, each frame $I_i$ is independently encoded using the frozen
DINOv2 ViT-S/14 encoder~\citep{dinov2}. We use its normalized
384-dimensional CLS representation
\[
\mathbf s_i\in\mathbb R^{384}
\]
as the frame descriptor. DINOv2 is the default descriptor used in the
main experiments; Table~\ref{tab:ablation-index} compares it with several
alternative place-recognition descriptors.

We maintain two mappings between the visual descriptor and the memory
address:
\[
W_{sh}\in\mathbb R^{1024\times384},
\qquad
W_{hs}\in\mathbb R^{384\times1024},
\]
such that
\[
W_{sh}\mathbf s_i\approx\mathbf h_i,
\qquad
W_{hs}\mathbf h_i\approx\mathbf s_i.
\]
The first converts appearance into an address, while the second provides
the reverse address-to-appearance association.

\paragraph{Why are these mappings learned online?}
Unlike the pose-to-address mapping, the association between a particular
visual observation and its address is not known before that frame is
encountered. New pairs $(\mathbf s_i,\mathbf h_i)$ therefore arrive
throughout the sequence. We update both visual associations incrementally
using recursive least squares (RLS), so that new observations can be
added without retraining either the DINOv2 encoder or the geometry
foundation model.

For a generic association
$W\mathbf x_i\approx\mathbf y_i$, the RLS update is
\begin{equation}
\begin{aligned}
\mathbf k_i
&=
\frac{P_{i-1}\mathbf x_i}
{1+\mathbf x_i^\top P_{i-1}\mathbf x_i},\\
W^{(i)}
&=
W^{(i-1)}
+
\left(
\mathbf y_i-W^{(i-1)}\mathbf x_i
\right)\mathbf k_i^\top,\\
P_i
&=
P_{i-1}
-
\mathbf k_i\mathbf x_i^\top P_{i-1},
\end{aligned}
\label{eq:app-rls}
\end{equation}
where $P_i$ is the RLS covariance matrix and $\mathbf k_i$ determines
the contribution of the new association.

For $W_{sh}$ we set
\[
\mathbf x_i=\mathbf s_i,
\qquad
\mathbf y_i=\mathbf h_i,
\]
while for $W_{hs}$ the two roles are exchanged. These updates modify
only the external memory.

\paragraph{The complete associative pathway.}
The four mappings together form
\[
\boxed{
\mathbf g_i
\;\xrightleftharpoons[W_{hg}]{W_{gh}}\;
\mathbf h_i
\;\xrightleftharpoons[W_{hs}]{W_{sh}}\;
\mathbf s_i
}
\]
and connect the spatial and visual information associated with a stored
frame. Here, $\mathbf g_i$ represents its global spatial state,
$\mathbf s_i$ represents its visual content, and $\mathbf h_i$ is the
shared sparse representation through which the two are associated.

The four matrices are obtained in different ways:
$W_{gh}$ is sampled once and remains fixed;
$W_{hg}$ is fitted once from sampled scaffold states;
and $W_{sh}$ and $W_{hs}$ are updated online as frames are stored.
None of these operations updates the geometry foundation model.

\paragraph{Retrieving a stored frame from a current observation.}
For a current frame $I_q$, we compute its descriptor $\mathbf s_q$ and
use only the appearance-to-address direction of the associative memory:
\[
\widetilde{\mathbf h}_q
=
\frac{W_{sh}\mathbf s_q}
{\|W_{sh}\mathbf s_q\|_2+10^{-12}}.
\]
We then compare this predicted address with every stored address,
\[
\operatorname{score}(q,j)
=
\widetilde{\mathbf h}_q^\top\mathbf h_j.
\]
A larger score means that the appearance of the current frame predicts
an address closer to the address stored for $I_j$.

We retain at most the five highest-scoring candidates and apply the
eligibility conditions from Sec.~\ref{sec:memory-retrieval}:
\begin{equation}
j<r_k-\Delta,
\qquad
\operatorname{src}(j)\neq k-1,
\qquad
\mathbf s_q^\top\mathbf s_j\geq\tau_{\mathrm{app}}.
\label{eq:app-retrieval-eligibility}
\end{equation}
The first condition prevents a temporally nearby frame from being treated
as a long-range revisit. The second removes the immediately preceding
source window, which is already connected to the current window through
their overlap. The third requires the retrieved frame to remain visually
compatible with the query. The surviving frames are only candidates;
their geometric consistency is checked in
Sec.~\ref{sec:retrieval-check} and expanded in
Appendix~\ref{app:closure}.

 Candidate
retrieval follows
\[
\mathbf s_q
\;\xrightarrow{W_{sh}}\;
\widetilde{\mathbf h}_q
\;\longrightarrow\;
\mathbf h_j
\;\longrightarrow\;
I_j,\operatorname{src}(j).
\]
In particular, we do not decode
$\widetilde{\mathbf h}_q$ through $W_{hg}$ and use the resulting state
as the current frame's global pose. Once a frame is retrieved, its source
window identifies its current global placement
$T_{\operatorname{src}(j)}$, while new camera predictions from the
geometry foundation model provide the relative transformation used to
verify the revisit.

\section{Scaffold Dynamics and Controlled Memory Diagnostics}
\label{app:association}

The previous section described the scaffold as the mapping
$\mathbf g\leftrightarrow\mathbf h\leftrightarrow\mathbf s$ used to
store and retrieve frames. Here we briefly describe the continuous
attractor structure underlying $\mathbf g$ and report controlled tests
of its behavior. These tests characterize the memory representation
itself; they are separate from the end-to-end trajectory experiments.

\paragraph{Continuous attractor fields.}
Each spatial or orientation component of the scaffold is represented by
a periodic neural field, following continuous-attractor models of spatial
and directional coding~\citep{burak2009accurate,bicanski2020neuronal}.
The field maintains a localized region of high activation, which we call
an \emph{activity bump}. The position of this bump around a ring or
torus represents the encoded coordinate.

For example, an orientation ring represents one angular variable such
as yaw. Moving the bump around the ring changes the represented angle,
while a full revolution returns to the same state. The spatial modules
use the same principle on periodic two-dimensional grids and height
rings. Thus, the scaffold stores pose through the \emph{location} of the
activity pattern rather than through its overall magnitude.

\paragraph{Maintaining and moving the represented pose.}
Local excitation together with broader inhibition allows an activity
bump to remain stable once formed~\citep{amari1977dynamics}. A motion
input can translate the bump across the periodic field, so changes in
camera position or orientation correspond to changes in its phase.
Observed camera poses provide anchors for the represented state; the
scaffold therefore does not rely only on accumulating motion over time.

The main \smr{} pipeline does not use these dynamics as a replacement
for the geometry foundation model's camera predictions. Instead, the
predicted global pose provides the state that is encoded when a frame is
stored, as described in Sec.~\ref{sec:memory-storage}.

\paragraph{Returning from an address to a valid scaffold state.}
Appendix~\ref{app:vhash} introduced the return mapping
$W_{hg}$ from the sparse address $\mathbf h$ to the scaffold state
$\mathbf g$. For a state $\mathbf g$, one complete scaffold return is
\[
\Phi(\mathbf g)
=
\psi\!\left(
W_{hg}\,
\sigma_{64}(W_{gh}\mathbf g)
\right),
\]
where $\psi$ normalizes the individual spatial and orientation fields.

This return mapping is useful because an associative cue need not
reproduce every component of the original scaffold state exactly.
The return operation encourages the reconstructed representation to
remain consistent with the learned scaffold structure. A perturbation
that makes different modules mutually inconsistent can therefore be
reduced by the return mapping. By contrast, a coherent displacement that
moves all modules to another valid pose is itself a valid scaffold state;
determining whether such a state corresponds to the correct physical
location requires the geometric verification described in
Sec.~\ref{sec:retrieval-check}.

\begin{figure}[t]
\centering
\includegraphics[width=\linewidth]{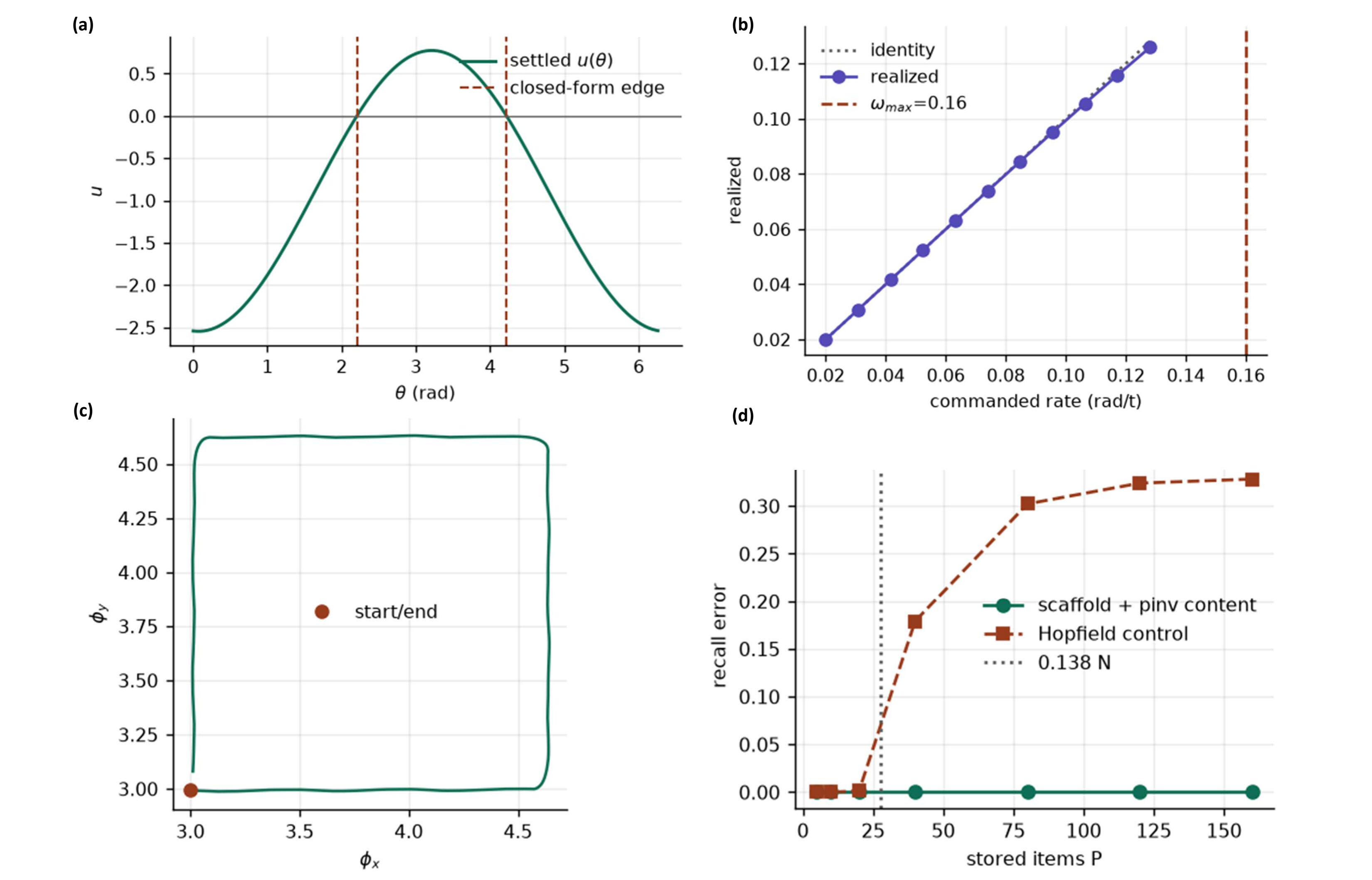}
\caption{
\textbf{Controlled diagnostics of the scaffold representation.}
\textbf{(a)} A stationary orientation ring maintains a localized
activity bump; the bump centre represents the encoded angle, while the
orange dashed lines show the predicted bump boundaries.
\textbf{(b)} When angular motion is supplied, the measured bump velocity
closely follows the commanded velocity over the tested range.
\textbf{(c)} The spatial phases during a closed two-dimensional path
return close to their starting point, illustrating path integration in
the periodic scaffold coordinates.
\textbf{(d)} Associative recall error as the number of stored states
increases. The scaffold configuration with the fitted content return map
maintains low error through the largest tested load ($P=160$), whereas
the Hopfield-style control accumulates larger error. These are controlled
memory diagnostics and do not measure end-to-end reconstruction accuracy
or establish an unrestricted memory-capacity limit.
}
\label{fig:scaffold-dyn}
\end{figure}

\paragraph{What these diagnostics establish.}
Fig.~\ref{fig:scaffold-dyn} checks three properties required by the
memory design: a pose representation can remain stable, its phase can
move continuously with motion, and a stored scaffold state can be
recalled through the associative representation over the tested memory
load. These experiments support the internal behavior of the scaffold,
while the usefulness of scaffold retrieval for \smr{} is evaluated
separately by the end-to-end comparison with flat descriptor retrieval
in Table~\ref{tab:ablation-index}.

\section{Geometric Verification of Retrieved Revisits}
\label{app:closure}

Section~\ref{sec:retrieval-check} describes how \smr{} verifies the
frames returned by the scaffold before changing any window placement.
Here we expand that procedure. The key idea is that a retrieved frame is
not trusted from appearance alone: it must provide a geometric connection
between the current window and a previously placed source window, and a
second source window must support a compatible placement.

\paragraph{Forming a retrieved pair.}
For each retrieved frame $I_j$, we select a nearby stored frame
$I_{j-3}$ or $I_{j+3}$. The partner must belong to the same source
window as $I_j$ and satisfy the sequential eligibility rule from
Sec.~\ref{sec:memory-retrieval}. If neither partner frame is valid, the
retrieved frame is discarded.

The two frames form a \emph{retrieved pair}. A pair is used instead of a
single frame because two camera poses provide a geometric baseline for
aligning the new prediction with the stored source window. We use an
offset of three frames as a fixed design choice so that the second frame
is nearby but distinct.

\paragraph{Why use two different source windows?}
We retain two retrieved pairs, denoted $a$ and $b$. The two frames within
each pair come from the same source window, but the two pairs must come
from different source windows. Because every stored frame records its
source window, we enforce$\operatorname{src}(j_a)\neq\operatorname{src}(j_b).$

This distinction is important. Each source window has its own global
placement $T$. If both pairs came from the same source window, both
placement proposals would depend on the same historical reference.
Using two different source windows instead provides two separately
placed references with which to check the current window.

We use two pairs by default. One pair provides no second historical
placement with which to test agreement, while a third pair requires two
additional frames. Table~\ref{tab:ablation-index} compares one, two, and three
pairs and shows that two retain most of the measured benefit of three
with fewer additional inputs.

\paragraph{Joint prediction with the current window.}
If two valid pairs are available, the frozen geometry foundation model
processes the 32 current-window frames together with the four retrieved
frames, giving 36 input frames in total. The 32 current frames retain
their sequential order, and the four retrieved frames are appended as
the two pairs.

Because all 36 frames are processed together, their camera poses are
predicted in one common local coordinate system. This joint prediction
creates a direct geometric relationship between the current window and
the retrieved historical frames; the existing chain of window placements
is not used to create that relationship.

\paragraph{Obtaining a global-placement proposal.}
Consider retrieved pair $a$, whose source window is
$\mathcal W_{o_a}$. The two frames in this pair have camera poses from
the new joint prediction and stored camera poses in the local coordinate
system of $\mathcal W_{o_a}$.

Using the similarity-alignment procedure from
Appendix~\ref{app:sim3}, we align these two sets of camera poses and
obtain
\[
\widehat Z_{o_a,k},
\]
the measured transformation relating the current joint prediction to
source window $\mathcal W_{o_a}$.

Since $\mathcal W_{o_a}$ already has the global placement $T_{o_a}$,
pair $a$ proposes the current window placement
\[
T_k^{(a)}
=
T_{o_a}\widehat Z_{o_a,k}.
\]
Repeating exactly the same procedure for pair $b$ gives
\[
T_k^{(b)}
=
T_{o_b}\widehat Z_{o_b,k}.
\]

Thus, the two retrieved pairs independently of the retrieval score provide
two geometric proposals for where the current window should lie in the
global reconstruction. The scaffold is used to find the historical
frames; the geometry foundation model and their stored source-window
placements determine these proposals.

\paragraph{Checking whether the two proposals agree.}
We accept the revisit only if $T_k^{(a)}$ and $T_k^{(b)}$ place the
current window sufficiently close to one another.

Let $E_k$ denote the median distance of the current-window camera
positions from their mean position after global placement. We use this
quantity to scale the translation tolerance to the size of the current
reconstruction. The two proposals must satisfy
\begin{equation}
\angle\!\left(
R(T_k^{(a)}),
R(T_k^{(b)})
\right)
\leq 10^\circ,
\qquad
\left\|
c(T_k^{(a)})-c(T_k^{(b)})
\right\|_2
\leq E_k .
\label{eq:site-agreement}
\end{equation}
The first condition checks rotational agreement, while the second checks
translation relative to the scale of the scene.

If either condition fails, the retrieved frames are rejected and no
retrieval-based placement correction is applied. If both conditions
pass, the revisit is accepted and the resulting geometric constraint is
used to revise the window placements, as described in
Appendix~\ref{app:pgo}.

\paragraph{What verification does and does not guarantee.}
The two source windows provide two separate geometric checks, but this
does not guarantee that every accepted revisit is correct. Repeated
scene structure, inaccurate historical placements, or weak two-camera
geometry can still produce compatible but incorrect proposals. The
agreement test therefore acts as a practical consistency check rather
than a proof of correspondence.

Jointly processing the retrieved frames can also change the current
window's local camera poses and point maps before the agreement decision
is made. Rejecting a revisit prevents the global placement correction,
but does not automatically restore the prediction obtained from the
current window alone. We examine this separate effect in
Appendix~\ref{app:anchored-distortion}. Our default configuration therefore retains the joint prediction only when it stays within $5^\circ$ of the prediction from the current window alone.

\section{Trajectory Revision and Global Refinement}
\label{app:pgo}

Section~\ref{sec:placement-update} describes the online correction applied
after a revisit has been verified, while
we also do the final refinement after the
full sequence has been processed. Both operations modify the global window
placements $T_k$; the camera poses and point maps stored inside each source
window remain in their local coordinate systems, as explained in
Appendix~\ref{app:invariant}.

\paragraph{From a verified revisit to a placement correction.}
Before a revisit, the current window $\mathcal W_k$ has global placement
$T_k$. The two retrieved pairs first produce the proposals
$T_k^{(a)}$ and $T_k^{(b)}$ described in
Appendix~\ref{app:closure}. These proposals are used only for the
agreement test in Eq.~\ref{eq:site-agreement}.

If the two proposals agree, we combine the frames from both retrieved
pairs and fit one similarity transformation using the same robust
alignment as Eq.~\ref{eq:sim3-alignment}. Specifically,
\begin{equation}
T_k^{\mathrm{corr}}
=
\argmin_{S\in\mathrm{Sim}(3)}
\left[
\sum_{j\in a}
w_j
\left\|
S\!\cdot c_j^{\mathcal A}
-
T_{o_a}\!\cdot c_j^{o_a}
\right\|_2^2
+
\sum_{j\in b}
w_j
\left\|
S\!\cdot c_j^{\mathcal A}
-
T_{o_b}\!\cdot c_j^{o_b}
\right\|_2^2
\right].
\label{eq:corrected-placement}
\end{equation}
Here, $S\in\mathrm{Sim}(3)$ is a candidate similarity transformation,
comprising scale, rotation, and translation, that maps the joint-pass
coordinate system into the global coordinate system. The weight $w_j$
is the robust weight assigned to retrieved frame $j$ by the alignment
procedure in Eq.~\ref{eq:sim3-alignment}, which downweights frames whose
camera poses are inconsistent with the fitted transformation.
Furthermore, $c_j^{\mathcal A}$ is the camera centre of retrieved frame
$j$ predicted in the joint pass $\mathcal A$, while $c_j^{o_a}$ or
$c_j^{o_b}$ is its stored camera centre in source window
$\mathcal W_{o_a}$ or $\mathcal W_{o_b}$, respectively. Applying
$T_{o_a}$ or $T_{o_b}$ places the corresponding stored camera centre
in the global coordinate system. Thus, $T_k^{\mathrm{corr}}$ is the
single robust placement that jointly aligns the predictions of both
retrieved pairs with their stored global poses; it is not an average of
$T_k^{(a)}$ and $T_k^{(b)}$.

The transformation required to move the current placement to this
corrected placement is
\begin{equation}
C
=
T_k^{\mathrm{corr}}T_k^{-1}.
\label{eq:placement-correction}
\end{equation}
Thus, $C$ represents the accumulated placement error revealed by the
revisit.

\paragraph{Why distribute the correction across earlier windows?}
The error revealed at window $k$ was generally accumulated over several
previous window alignments. Applying the full correction only to
$\mathcal W_k$ would therefore create a discontinuous jump between
neighboring windows. Instead, \smr{} distributes the correction over the
windows between the earlier reference window $\mathcal W_o$ and the
current window.

For a window $\mathcal W_\ell$ with $o<\ell\leq k$, we define
\[
\alpha_\ell
=
\frac{\ell-o}{k-o}.
\]
The reference window has correction weight $0$, while the current window
has weight $1$. Intermediate windows receive progressively larger
fractions of the correction.

\paragraph{Applying a fraction of a similarity transformation.}
A similarity transformation contains rotation, translation, and scale,
so it cannot be scaled directly by multiplying its transformation matrix
by $\alpha_\ell$. We instead use the logarithm and exponential maps:
\begin{equation}
T_\ell
\leftarrow
\operatorname{Exp}\!\left(
\alpha_\ell
\operatorname{Log}C
\right)
T_\ell,
\qquad
o<\ell\leq k.
\label{eq:app-placement-update}
\end{equation}
Here,
\[
\operatorname{Log}C
=
(\boldsymbol\omega,\boldsymbol\nu,\sigma)
\in\mathbb R^7
\]
represents the correction using three rotation values, three
translation-related values, and one log-scale value.
Multiplying this vector by $\alpha_\ell$ selects the desired fraction of
the correction, and $\operatorname{Exp}$ converts it back into a valid
similarity transformation.

Consequently,
\[
\alpha_\ell=0
\quad\Rightarrow\quad
\text{no correction},
\]
while
\[
\alpha_\ell=1
\quad\Rightarrow\quad
\text{the full correction }C.
\]
The update therefore changes the trajectory smoothly between the
historical reference and the current window rather than inserting a
single abrupt correction.

\paragraph{What changes after the update?}
Only the global placements $T_\ell$ are modified. The camera poses and
point maps stored for frames in $\mathcal W_\ell$ remain unchanged in
their source-window coordinates. Applying the updated $T_\ell$ therefore
moves the cameras and their associated 3D geometry together. This is the
same separation between local geometry and global placement described in
Appendix~\ref{app:invariant}.

The online update is immediately available to later windows. Therefore,
a correction made at one revisit can influence the global placements
used when subsequent frames are stored and when later revisits are
verified.

\paragraph{Recording constraints between windows.}
In addition to updating the running trajectory, \smr{} retains the
relative geometric measurements obtained while processing the sequence.
There are two types of connections.

A \emph{sequential connection} is obtained from the shared frames of two
neighboring windows. A \emph{revisit connection} is obtained from an
accepted historical retrieval. For a connection
$e=(\ell,k)$, let $Z_e$ denote the measured relative transformation
between windows $\mathcal W_\ell$ and $\mathcal W_k$.

The current global placements themselves imply the relative
transformation
\[
T_\ell^{-1}T_k.
\]
If the reconstruction is perfectly consistent with the measurement
$Z_e$, these two relative transformations should agree.

\paragraph{Final global refinement.}
Online corrections are applied one revisit at a time. After all windows
have been processed, \smr{} jointly refines the complete set of window
placements using all retained sequential and accepted revisit
connections.

We fix $T_1$ to remove the global gauge freedom and optimize
\begin{equation}
\min_{T_2,\ldots,T_K}
\sum_{e=(\ell,k)\in\mathcal E}
\rho_H\!\left(
\left\|
\operatorname{Log}
\left(
Z_e^{-1}T_\ell^{-1}T_k
\right)
\right\|_{\Lambda_e}^{2}
\right),
\label{eq:app-pose-graph}
\end{equation}
where $\mathcal E$ is the set of retained connections.
The quantity
\[
Z_e^{-1}T_\ell^{-1}T_k
\]
measures the disagreement between the relative placement predicted by
the current trajectory and the relative transformation measured from
the frames. Taking $\operatorname{Log}$ converts this disagreement into
rotation, translation, and scale components that can be optimized
together.

The diagonal matrix $\Lambda_e$ controls the relative contribution of
these components, while the Huber loss $\rho_H$~\citep{huber} reduces
the influence of unusually large residuals. This is a pose-graph
optimization over window placements rather than bundle adjustment over
individual 3D points~\citep{grisetti,kummerle2011g,dellaert2012factor}.

\paragraph{Handling inconsistent revisit constraints.}
An accepted revisit can still be incorrect, for example, because of
repeated scene structure. We therefore perform the global solve
robustly. After the first optimization, revisit connections with large
residuals are rejected according to the configured residual criterion,
and the remaining graph is optimized once more. Sequential connections
are retained because they define the basic chain of the reconstruction.

\paragraph{Online revision versus final refinement.}
The two stages serve different purposes. The distributed update in
Eq.~\ref{eq:app-placement-update} corrects the trajectory while the
sequence is still being processed, so later observations benefit from
the revised placements. The optimization in
Eq.~\ref{eq:app-pose-graph} instead considers all measurements together
after the sequence is complete and removes remaining global
inconsistencies. Both stages update only the window placements.

\section{Datasets and Evaluation Protocols}
\label{app:datasets}

The main experiments in Sec.~\ref{sec:experiments} evaluate \smr{} on
five downstream tasks using five datasets. Here we provide the exact
evaluation subsets, frame-selection rules, and scoring procedures used
for each task. Within every paired Baseline/$+\smr{}$ comparison, the
same geometry foundation model and the same input frames are used.
When computational constraints require a reduced subset for a particular
model, its Baseline and $+\smr{}$ results are evaluated on the same
subset.

\begin{table}[t]
\centering
\scriptsize
\setlength{\tabcolsep}{1.2pt}
\caption{
\textbf{Datasets and evaluation protocols.}
The table summarizes the data used for each downstream task. Detailed
sampling and scoring procedures are given below.
}
\label{tab:datasets}

\begin{tabular}{@{}
>{\raggedright\arraybackslash}p{0.22\columnwidth}
>{\raggedright\arraybackslash}p{0.24\columnwidth}
>{\raggedright\arraybackslash}p{0.47\columnwidth}
@{}}
\toprule
\textbf{Task} & \textbf{Dataset} & \textbf{Evaluation subset} \\
\midrule

Camera pose
& CO3Dv2~\citep{co3d}
& 37 validated trajectories;
$N\in\{10,50,100,200\}$ sampled frames \\

Camera pose
& RE10K~\citep{re10k}
& 1,653 clips at $N=10$; 100 clips for the
$N\in\{50,100,200\}$ length study \\

Multi-view depth
& Replica~\citep{replica}
& 18 scenes; $N=2000$ frames per scene \\

3D point-cloud reconstruction
& ScanNet++~\citep{scannetpp}
& $1,006$ scenes; $N=600-8,000$ frames per scene \\

SLAM
& 7-Scenes~\citep{sevenscenes}
& All seven indoor scenes \\

Novel view synthesis
& 7-Scenes~\citep{sevenscenes}
& Seven seq-01 trajectories, 200 keyframes each; every eighth keyframe held out (156 targets); decoders trained on the other 7-Scenes sequences \\

Novel view synthesis
& CO3Dv2~\citep{co3d}
& 37 validated trajectories, 200 frames each; every eighth frame held out (888 targets); test only, no adaptation \\

\bottomrule
\end{tabular}
\end{table}

\subsection{Camera Pose Estimation}
\label{app:pose-eval}

The camera-pose experiments in Sec.~\ref{sec:results-pose} evaluate how
pose accuracy changes as more frames from the same video trajectory are
processed. We use CO3Dv2 and RealEstate10K because they provide different
camera-motion patterns: CO3Dv2 contains object-centric trajectories with
substantial viewpoint changes, whereas RE10K is dominated by forward
camera motion and therefore contains fewer natural revisits.

\paragraph{CO3Dv2.}
We use 37 CO3Dv2 sequences with validated camera trajectories. The videos
span approximately 8--68\,s, with a median duration of 26\,s. For each
trajectory, we evaluate
\[
N\in\{10,50,100,200\}
\]
input frames. Frames are sampled uniformly across the available
trajectory rather than taking the first $N$ frames. Thus, increasing
$N$ increases the temporal density with which the same underlying
trajectory is observed.

The controlled pose ablations in
Appendix~\ref{app:ablations} use a separate subset of twelve CO3Dv2
trajectories at $N=200$ and should not be interpreted as the full
37-sequence evaluation.

\paragraph{RealEstate10K.}
RE10K contains real-estate videos with predominantly forward camera
motion~\citep{re10k}. The 100 clips used for the length study span
6.7--9.3\,s, with a median duration of 9.3\,s. We uniformly sample
$N\in\{50,100,200\}$ frames from these clips. For the $N=10$
evaluation, we use the larger set of 1,653 usable test clips.

The $N=10$ and longer-sequence evaluations therefore use different
numbers of clips and should not be interpreted as a nested set of the
same sequences.

\paragraph{Pose metric.}
Following the relative-pose evaluation used in the main experiments, we
evaluate every pair of sampled frames. For a pair $(i,j)$, we compute
its rotation error and translation-direction error and define the
pairwise pose error as
\[
e_{ij}
=
\max
\left(
e_{ij}^{\mathrm{rot}},
e_{ij}^{\mathrm{trans}}
\right).
\]
Translation magnitude is not scored because monocular geometry is defined
only up to scale.

AUC@30 is the normalized area under the cumulative distribution of
$e_{ij}$ between $0^\circ$ and $30^\circ$. Higher AUC therefore means
that a larger fraction of frame pairs have accurate relative camera
poses. Table~\ref{tab:pose} reports the mean and standard deviation over
the evaluated sequences or clips.

\subsection{Multi-view Depth Estimation and 3D Point-Cloud Reconstruction}
\label{app:dense-eval}

The experiments in Sec.~\ref{sec:mvs} evaluate multi-view depth
estimation on Replica~\citep{replica} and 3D point-cloud reconstruction
on ScanNet++~\citep{scannetpp}. Both are evaluated in the same
long-horizon setting using 32-frame windows with 16-frame overlap, as
described in Sec.~\ref{sec:method}. Each window is
first reconstructed locally by the frozen geometry foundation model and
then placed in the shared global coordinate system using the overlapping
frames. Baseline and $+\smr{}$ use exactly the same input frames,
windowing strategy, and geometry foundation model. Baseline connects the
windows sequentially without long-range memory, whereas $+\smr{}$
additionally retrieves earlier observations, verifies candidate revisits,
and uses accepted geometric evidence to revise the global placements of
the reconstructed windows.

\paragraph{Replica.}
Replica~\citep{replica} provides photorealistic synthetic indoor
environments together with accurate camera information and dense
reference 3D geometry. It contains 18 such scenes with about 2000 frames per scene. Its controlled geometry allows us to measure how
reconstruction errors accumulate as a scene is processed across many
overlapping windows. For each evaluated trajectory, Baseline and
$+\smr{}$ process the same frames in the same temporal order; the only
difference is whether the external memory is used to retrieve and revise
previously placed reconstruction windows.

\paragraph{ScanNet++.}
ScanNet++~\citep{scannetpp} contains high-resolution observations of
real indoor environments together with high-quality reference 3D
geometry. It contains 1006 scenes with about 8000 frames per scene. Compared with Replica, it evaluates the same long-horizon
reconstruction procedure under real image appearance, camera motion, and
more complex scene geometry. We again use identical input frames and
32-frame windows for Baseline and $+\smr{}$, so improvements measure the
effect of memory-based revisit verification and trajectory revision
rather than differences in the input data or geometry foundation model.

\paragraph{Dense-geometry metrics.}
For both datasets, we report \emph{accuracy}, \emph{completeness}, and
\emph{overall} reconstruction error. Accuracy measures the distance from
the reconstructed geometry to the reference geometry, while completeness
measures the reverse distance from the reference geometry to the
reconstruction. We define the overall error as the mean of accuracy and
completeness. All distances are reported in centimetres, and lower values
indicate better reconstruction. Table~\ref{tab:pointcloud} reports the
mean and standard deviation over the corresponding evaluation set.

\subsection{Camera Pose Results Across Sequence Lengths}
\label{app:pose-results}

The full numerical results corresponding to
Fig.~\ref{fig:camera_pose} are reported in
Table~\ref{tab:pose}. The table evaluates the same Baseline/$+\smr{}$
pairs described in Appendix~\ref{app:pose-eval} as the number of input
frames increases from $N=10$ to $N=200$.

For the feed-forward geometry foundation models, the two methods are
identical at the shortest sequence lengths whenever the sequence fits
within the initial processing context. Differences become more apparent
as the sequence grows and predictions must be connected across multiple
windows. This pattern is strongest on CO3Dv2: at $N=200$, for example,
VGGT improves from 90.0 to 93.9 AUC@30, VGGT-$\Omega$ from 91.6 to
94.7, and $\pi^3$ from 89.9 to 93.3. The improvements on RE10K are
generally smaller, consistent with its predominantly forward camera
motion providing fewer opportunities to retrieve previously observed
regions.

The streaming models expose a complementary effect. Their native causal
inference becomes increasingly expensive with sequence length and exceeds
the tested 46\,GB memory budget for several longer inputs. The
windowed \smr{} configuration remains executable at these lengths while
retaining access to earlier observations through the external memory.
These comparisons use the same sampled frames whenever both variants can
be evaluated.

\begin{table*}[t]
\centering
\scriptsize
\setlength{\tabcolsep}{0.25pt}
\renewcommand{\arraystretch}{1.10}
\caption{
\textbf{Camera pose estimation accuracy on RE10K~\citep{re10k} and
CO3Dv2~\citep{co3d}.}
We report AUC@30$\uparrow$ for increasing numbers of input frames $N$.
Values are mean $\pm$ standard deviation across evaluated sequences
(CO3Dv2) or clips (RE10K).
For each geometry foundation model, Baseline denotes its original
inference procedure and $+\smr{}$ denotes the same model equipped with
our framework. For StreamVGGT and STream3R, Baseline uses native causal
inference, whereas $+\smr{}$ uses our windowed framework.
OOM indicates that inference exceeded the tested 46\,GB GPU memory
budget. Within each model and sequence length, the higher mean is shown
in bold; equal means are left unbolded.
}
\label{tab:pose}

\begin{tabular*}{\textwidth}{
@{}c l l @{\extracolsep{\fill}} cccc @{\hspace{2pt}} cccc@{}}
\toprule

& \textbf{Model} & \textbf{Method}
& \multicolumn{4}{c}{\textbf{RE10K} ($N$)}
& \multicolumn{4}{c}{\textbf{CO3Dv2} ($N$)}
\\
\cmidrule(lr){4-7}
\cmidrule(lr){8-11}

& &
& 10 & 50 & 100 & 200
& 10 & 50 & 100 & 200
\\
\midrule

\multirow{12}{*}{\rotatebox[origin=c]{90}{\textit{Feed-forward}}}

& DUSt3R & Baseline
& 65.6$\pm$2.8
& 88.0$\pm$0.7
& \textbf{85.9$\pm$0.7}
& 82.6$\pm$0.9
& 88.9$\pm$1.3
& 88.8$\pm$0.7
& 88.0$\pm$0.6
& 83.9$\pm$0.8
\\

& & $+\smr{}$
& 65.6$\pm$2.8
& 88.0$\pm$0.7
& 85.6$\pm$0.8
& \textbf{83.5$\pm$0.9}
& 88.9$\pm$1.3
& 88.8$\pm$0.7
& \textbf{89.5$\pm$0.5}
& \textbf{86.6$\pm$0.7}
\\
\addlinespace[1pt]

& Fast3R & Baseline
& 71.2$\pm$2.5
& 84.3$\pm$1.5
& 86.1$\pm$1.3
& \textbf{86.6$\pm$1.3}
& 90.9$\pm$1.6
& 88.3$\pm$1.6
& 86.7$\pm$1.3
& 82.4$\pm$1.3
\\

& & $+\smr{}$
& 71.2$\pm$2.5
& 84.3$\pm$1.5
& \textbf{87.0$\pm$1.4}
& 86.5$\pm$1.4
& 90.9$\pm$1.6
& 88.3$\pm$1.6
& \textbf{87.1$\pm$1.3}
& \textbf{84.3$\pm$1.3}
\\
\addlinespace[1pt]

& MASt3R & Baseline
& 85.5$\pm$2.1
& 95.3$\pm$0.2
& 95.2$\pm$0.3
& 93.8$\pm$0.3
& 93.0$\pm$1.3
& 89.6$\pm$1.5
& 88.5$\pm$1.3
& 86.6$\pm$1.1
\\

& & $+\smr{}$
& 85.5$\pm$2.1
& 95.3$\pm$0.2
& \textbf{95.3$\pm$0.3}
& \textbf{94.3$\pm$0.3}
& 93.0$\pm$1.3
& 89.6$\pm$1.5
& \textbf{88.7$\pm$1.3}
& \textbf{86.8$\pm$1.2}
\\
\addlinespace[1pt]

& VGGT & Baseline
& 88.7$\pm$2.6
& 88.5$\pm$1.9
& 88.7$\pm$1.8
& 88.5$\pm$1.7
& 95.7$\pm$0.8
& 94.5$\pm$0.9
& 92.2$\pm$1.1
& 90.0$\pm$0.7
\\

& & $+\smr{}$
& 88.7$\pm$2.6
& 88.5$\pm$1.9
& \textbf{89.6$\pm$1.8}
& \textbf{89.1$\pm$1.8}
& 95.7$\pm$0.8
& 94.5$\pm$0.9
& \textbf{92.6$\pm$1.2}
& \textbf{93.9$\pm$0.6}
\\
\addlinespace[1pt]

& VGGT-$\Omega$ & Baseline
& 88.9$\pm$2.0
& 95.7$\pm$1.1
& 95.4$\pm$1.1
& 94.9$\pm$1.1
& 95.7$\pm$0.9
& 93.2$\pm$1.3
& 93.6$\pm$0.6
& 91.6$\pm$0.6
\\

& & $+\smr{}$
& 88.9$\pm$2.0
& 95.7$\pm$1.1
& \textbf{95.5$\pm$1.1}
& \textbf{95.2$\pm$1.1}
& 95.7$\pm$0.9
& 93.2$\pm$1.3
& \textbf{94.2$\pm$0.7}
& \textbf{94.7$\pm$0.5}
\\
\addlinespace[1pt]

& $\pi^3$ & Baseline
& 88.2$\pm$2.1
& 96.4$\pm$0.8
& 95.9$\pm$1.0
& 95.2$\pm$1.1
& 96.3$\pm$0.7
& 94.2$\pm$1.3
& 93.7$\pm$0.6
& 89.9$\pm$0.7
\\

& & $+\smr{}$
& 88.2$\pm$2.1
& 96.4$\pm$0.8
& 95.9$\pm$1.0
& \textbf{95.4$\pm$1.0}
& 96.3$\pm$0.7
& 94.2$\pm$1.3
& \textbf{94.8$\pm$0.6}
& \textbf{93.3$\pm$0.7}
\\

\midrule

\multirow{4}{*}{\rotatebox[origin=c]{90}{\textit{Streaming}}}

& StreamVGGT & Baseline
& 81.2$\pm$0.6
& 78.6$\pm$0.5
& 77.7$\pm$0.6
& OOM
& 92.7$\pm$1.2
& 90.2$\pm$0.8
& 87.5$\pm$0.8
& OOM
\\

& & $+\smr{}$
& \textbf{86.7$\pm$0.6}
& \textbf{86.6$\pm$0.5}
& \textbf{86.8$\pm$0.6}
& \textbf{85.7$\pm$0.7}
& 92.7$\pm$1.2
& \textbf{92.8$\pm$0.8}
& \textbf{92.0$\pm$0.8}
& \textbf{85.9$\pm$1.0}
\\
\addlinespace[1pt]

& STream3R & Baseline
& 89.2$\pm$0.2
& 87.0$\pm$0.2
& OOM
& OOM
& 92.1$\pm$1.0
& OOM
& OOM
& OOM
\\

& & $+\smr{}$
& \textbf{93.8$\pm$0.2}
& \textbf{93.2$\pm$0.2}
& \textbf{92.7$\pm$0.4}
& \textbf{89.3$\pm$0.6}
& 92.1$\pm$1.0
& \textbf{89.0$\pm$0.9}
& \textbf{87.6$\pm$1.1}
& \textbf{87.6$\pm$0.8}
\\

\bottomrule
\end{tabular*}
\end{table*}

\subsection{Full 7-Scenes SLAM Results}
\label{app:slam-full}

Table~\ref{tab:slam-full} reports the full per-scene SLAM results for
7-Scenes. These are the detailed numbers summarized in the main-text
SLAM comparison. We report absolute trajectory error (ATE) RMSE in
metres, after similarity alignment to the ground-truth trajectory, so
lower is better.

The table separates calibrated and uncalibrated methods because these
settings are not directly comparable: calibrated methods receive camera
intrinsics, whereas uncalibrated methods do not. Within the
VGGT-$\Omega$ block, the \smr{} rows show the effect of online revision
and the final pose-graph refinement on the same frozen geometry
foundation model.

\begin{table}[t]
\centering
\scriptsize
\setlength{\tabcolsep}{1.2pt}
\caption{\textbf{Full 7-Scenes SLAM results.} We report per-scene and average ATE RMSE (m$\downarrow$) on the seven 7-Scenes sequences: chess, fire, heads, office, pumpkin, kitchen, and stairs. Lower is better. We compare NICER-SLAM, DROID-SLAM, MASt3R-SLAM, GLUEMAP, TALO, VGGT-SLAM++, ViSTA-SLAM, SLAM-Former, CUT3R, VGGT-SLAM, and VGGSfM~\citep{nicerslam,droidslam,mast3rslam,gluemap,talo,vggtslampp, vistaslam,slamformer,cut3r,vggtslam,vggsfm}. Calibrated methods receive camera intrinsics, whereas uncalibrated methods do not. A superscript $^{*}$ marks an uncalibrated version of a method that is also reported in a calibrated setting. Our results use frozen VGGT-$\Omega$ with 32-frame windows. Within each calibration block, the lowest value in each scene column is shown in bold.}
\label{tab:slam-full}

\begin{tabular*}{\linewidth}{
@{}l l @{\extracolsep{\fill}} cccccccc@{}}
\toprule
Model & Method
& chess & fire & heads & office & pumpkin & kitchen & stairs & avg \\
\midrule

\multicolumn{10}{@{}l}{\textit{Calibrated methods}}\\

NICER-SLAM
& --
& 0.033 & 0.069 & 0.042 & 0.108 & 0.200 & 0.039 & 0.108 & 0.086 \\

DROID-SLAM
& --
& 0.036 & 0.027 & 0.025 & 0.066 & 0.127 & 0.040 & 0.026 & 0.049 \\

MASt3R-SLAM
& --
& 0.053 & 0.025 & 0.015 & 0.097 & 0.088 & 0.041 & \textbf{0.011} & 0.047 \\

GLUEMAP
& --
& 0.036 & 0.031 & 0.018 & 0.079 & 0.092 & 0.046 & 0.051 & 0.050 \\

TALO
& --
& 0.034 & 0.021 & 0.012 & 0.054 & 0.061 & 0.028 & \textbf{0.010} & \textbf{0.031} \\

\midrule
\multicolumn{10}{@{}l}{\textit{Uncalibrated methods}}\\

VGGT-SLAM++
& --
& 0.034 & \textbf{0.023} & 0.017 & 0.104 & 0.127 & 0.085 & 0.060 & 0.064 \\

ViSTA-SLAM
& --
& 0.036 & 0.033 & 0.019 & 0.103 & 0.133 & 0.058 & 0.093 & 0.067 \\

SLAM-Former
& --
& 0.039 & 0.033 & 0.018 & 0.070 & 0.065 & 0.035 & 0.033 & 0.042 \\

CUT3R
& --
& 0.046 & 0.043 & 0.055 & 0.120 & 0.096 & 0.061 & 0.086 & 0.073 \\

MASt3R-SLAM$^{*}$
& --
& 0.063 & 0.046 & 0.029 & 0.103 & 0.114 & 0.074 & 0.032 & 0.066 \\

VGGT-SLAM
& Sim(3), $w{=}32$
& 0.037 & 0.026 & 0.018 & 0.104 & 0.133 & 0.061 & 0.093 & 0.067 \\

VGGT-SLAM
& SL(4), $w{=}32$
& 0.036 & 0.028 & 0.018 & 0.103 & 0.133 & 0.058 & 0.093 & 0.067 \\

VGGSfM
& --
& 0.052 & 0.038 & 0.021 & 0.076 & 0.104 & 0.059 & 0.061 & 0.059 \\

GLUEMAP$^{*}$
& --
& 0.042 & 0.039 & 0.022 & 0.088 & 0.105 & 0.054 & 0.059 & 0.058 \\

TALO$^{*}$
& --
& 0.028 & \textbf{0.023} & 0.014 & 0.049 & 0.082 & 0.033 & 0.038 & 0.038 \\

\midrule
\multicolumn{10}{@{}l}{\textit{VGGT-$\Omega$ (uncalibrated)}}\\

VGGT-$\Omega$
& Baseline
& 0.025 & 0.024 & \textbf{0.012} & 0.047 & 0.161 & 0.026 & 0.020 & 0.045 \\

VGGT-$\Omega$
& +\smr{}
& \textbf{0.021} & 0.031 & 0.013 & 0.031
& \textbf{0.061} & \textbf{0.025} & 0.022 & \textbf{0.029} \\

\bottomrule
\end{tabular*}
\end{table}

\subsection{Novel View Synthesis}
\label{app:nvs-data}
The novel view synthesis experiment in Sec.~\ref{sec:nvs} asks whether
the map maintained by \smr{} yields better rendered views than the chained
map of the same frozen geometry model. The test sequences are the seven
7-Scenes seq-01 trajectories of Sec.~\ref{sec:results-slam} (every fifth
frame is a keyframe, 200 keyframes per trajectory) and the 37 validated
CO3Dv2 trajectories of Sec.~\ref{sec:results-pose} (200 frames each).
Every eighth keyframe is a held-out target that is never shown to the
geometry model; the remaining keyframes are processed with the default
configuration of Table~\ref{tab:hparams} (32-frame windows with 16-frame
overlap, retrieval, verification, and online revision; no final global
refinement). The evaluated set contains 156 held-out targets on
7-Scenes and 888 on CO3Dv2 per geometry model. Baseline and $+\smr{}$
render the same targets from identical source keyframes using the matched
decoder-training procedure in Appendix~\ref{app:nvs}. CO3Dv2
is a test-only collection: the decoders are trained on 7-Scenes sequences
and applied to CO3Dv2 without adaptation. The complete rendering and
decoder-training protocol is given in Appendix~\ref{app:nvs}. We report PSNR, SSIM, and LPIPS~\citep{lpips}.

\begin{figure}[t]
\centering
\includegraphics[width=\linewidth]{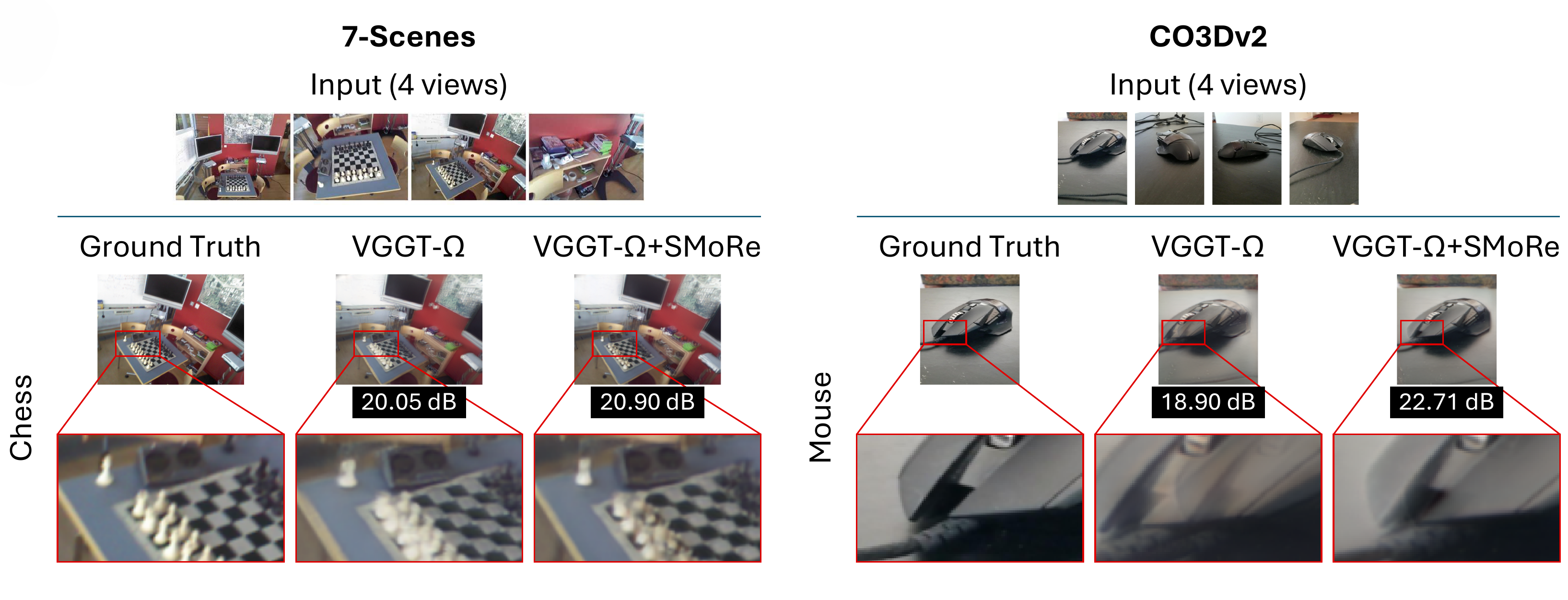}
\caption{
\textbf{Additional novel view synthesis results on 7-Scenes and CO3Dv2.}
On 7-Scenes \emph{chess} (left), \smr{} suppresses ghosting artifacts around the chess pieces, yielding sharper boundaries. On CO3Dv2 (right), the zoomed-in views of the computer mouse demonstrate that \smr{} mitigates edge misalignment and better preserves object contours. Together, these examples highlight consistent improvements in boundary sharpness and geometric structure across diverse scenes. In both panels, columns from left to right display the ground truth, the baseline VGGT-$\Omega$ rendering, and our results with \smr{}. Above each comparison group, we show four representative input keyframes out of the eight source views used to render each target. Red boxes indicate the cropped regions shown in the magnified insets.
}
\label{fig:nvs-extra-viz}
\end{figure}

\paragraph{Qualitative results.}
Fig.~\ref{fig:nvs-extra-viz} extends the comparison in Fig.~\ref{fig:camera_pose}B to additional scenes across datasets. We compare VGGT-$Omega$ with and without \smr{} on representative scenes from 7-Scenes (left) and Co3Dv2 (right). The enlarged regions highlight sharper object boundaries and better-preserved scene details with \smr{}, particularly around the chess pieces (left) and computer mouse (right). These improvements suggest that revising window placements brings observations of the same surfaces into closer geometric alignment, consistent with the quantitative gains reported in Table~\ref{tab:nvs}.

\FloatBarrier

\section{Novel View Synthesis Protocol and Decoder}
\label{app:nvs}
We evaluate whether revising stored geometry improves novel view
synthesis from long sequences. We use the same frozen geometry models,
windowing scheme, and memory configuration as in the camera pose
experiments. A Gaussian decoder converts the stored geometry into
rendered images. The datasets and evaluation splits are described in
Appendix~\ref{app:nvs-data}.

\subsection{Source Views and Geometry}
\label{app:nvs-cameras}
We render each held-out target from eight nearby input keyframes.
Source views are selected using the sum of camera-centre distance and
viewing-direction angle, computed from the ground-truth cameras.
The target itself is excluded. Baseline and $+\smr{}$ use identical
source views.

Each source keyframe retains the depth, confidence, and camera intrinsics
predicted when it first enters a window. Its depth is unprojected into
3D points in that window's local coordinate system
(Appendix~\ref{app:invariant}). Baseline places these points in the global
map using chained window transformations $T_k$. \smr{} uses the window
transformations obtained after online revision. We fit a similarity
transformation between the estimated and ground-truth input cameras.
We use this alignment to express the target camera in each map's
coordinate system.

Before decoding, we centre the source points at the centre of the source
cameras and normalize their scale by the median source depth. We discard
the lowest-confidence 30\% of points in each frame. We also remove points
at depth discontinuities, defined by a relative depth difference above
5\% between neighbouring pixels. Both systems use the same filtering
procedure. Images are rendered at $384\times288$ resolution with a 4:3
aspect ratio. Portrait CO3Dv2 sequences retain their original orientation.

\subsection{Gaussian Decoder and Training}
\label{app:nvs-training}
We use a Gaussian decoder to predict rendering parameters from the
source images and their geometry. Each source pixel provides eleven
input channels:
\[
\underbrace{3}_{\text{RGB}}
+\underbrace{3}_{\text{map position}}
+\underbrace{1}_{\text{depth}}
+\underbrace{1}_{\text{log-confidence}}
+\underbrace{3}_{\text{ray direction}}
=11.
\]
A four-level U-Net with approximately 7M parameters predicts a position
offset, log-scales, rotation, opacity, and colour residual for each valid
pixel. The position offset is bounded by two pixel footprints, and the
scales are predicted relative to the footprint. A pixel's footprint is
its depth multiplied by the pixel size and divided by the focal length.
This parameterization scales the Gaussians according to the input geometry.

The Gaussians from all source views are rasterized into the target view
using gsplat~\citep{kerbl3dgs,gsplat}. An image-space U-Net with
approximately 3M parameters then refines the rendering. It takes the
rendered colour, alpha, and depth as input and predicts a bounded colour
residual. The decoder is initialized to approximate splatting the coloured
source points.

\paragraph{Training data.}
For each geometry model, we train one decoder on the chained geometry
and another on the geometry maintained by \smr{}. Both decoders use the
same architecture, training data, schedule, and random seed. Training
uses all 7-Scenes sequences except seq-01. These sequences are processed
with the same windowing and memory configuration as the test sequences,
and every keyframe is stored. Each training sample consists of one target
keyframe and its eight nearest source keyframes. We exclude the target
and its immediate neighbours from the source set, so the target image
does not provide source pixels.

\paragraph{Optimization.}
We train the decoders with an image reconstruction loss:
\[
\mathcal L=\mathcal L_{1}+0.5\,\mathcal L_{\mathrm{LPIPS}}+0.2\,(1-\mathrm{SSIM}).
\]
We use AdamW for 12K steps with a learning rate of $2\times10^{-4}$,
500 warm-up steps, and cosine decay. Gradients are accumulated over two
samples. Training takes approximately one hour per decoder on an RTX
6000 Ada GPU. Neither the geometry foundation model nor \smr{} is
optimized by this loss. The comparison therefore evaluates each geometry
representation with its own trained decoder under matched training
conditions.

\subsection{Evaluation and Additional Results}
\begin{figure}[t]
\centering
\includegraphics[width=\linewidth]{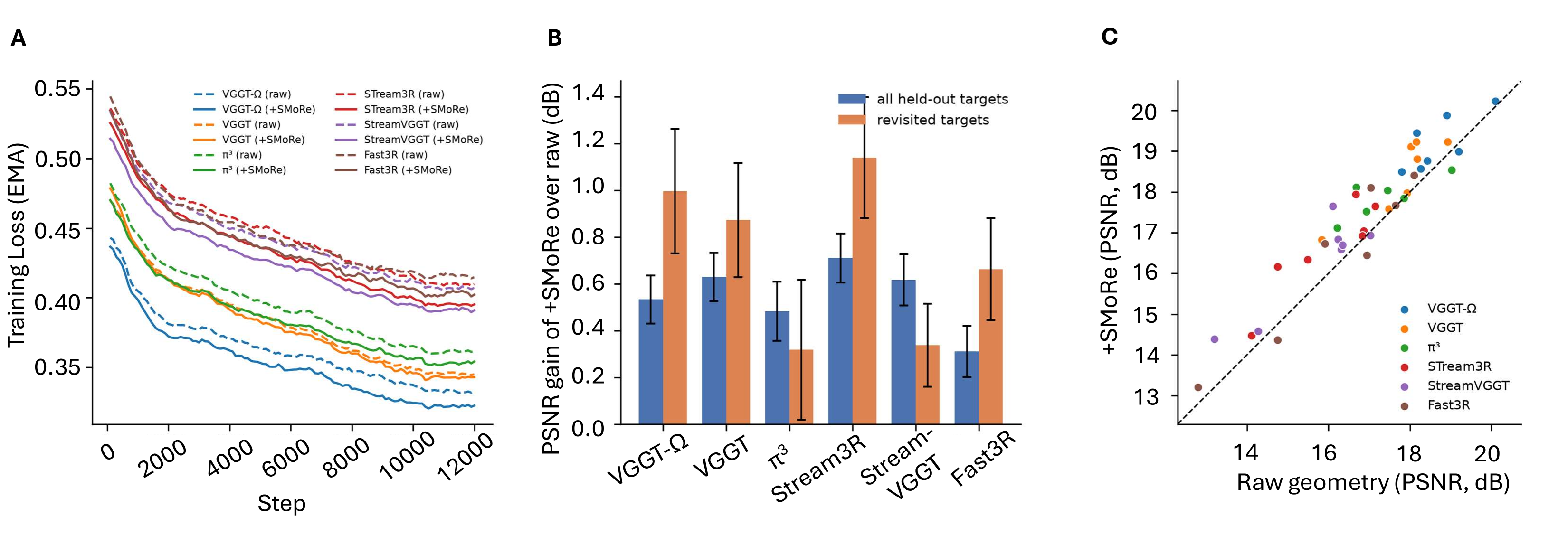}
\caption{
\textbf{Gaussian decoder training and novel view synthesis results.}
\textbf{(A)} Training loss, shown as an exponential moving average, for decoders
trained on chained geometry (dashed) and geometry maintained by \smr{}
(solid), for all six geometry models.
\textbf{(B)} Paired PSNR gains on 7-Scenes, reported as mean $\pm$ standard error
over all held-out targets and over revisited targets.
\textbf{(C)} Mean PSNR for each of the seven 7-Scenes trajectories, comparing
Baseline with $+\smr{}$. Points above the diagonal indicate improvement.
}
\label{fig:nvs-training}
\end{figure}

We render each held-out target from the chained map with the Baseline
decoder and from the revised map with the $+\smr{}$ decoder. We compute
PSNR, SSIM, and LPIPS against the target image, following
Appendix~\ref{app:nvs-data}. Gains are computed from paired differences
between the two renderings of each target.
Figure~\ref{fig:nvs-training} reports the paired PSNR gains on 7-Scenes
and their standard errors. Decoders trained on 7-Scenes are evaluated on CO3Dv2
without adaptation, testing transfer from indoor scenes to object-centric
sequences.

\paragraph{Results on revisited regions.}
On 7-Scenes, the mean paired PSNR gains range from 0.31 to 0.71\,dB
across the six geometry models, with standard errors of 0.10 to 0.13\,dB.
All six 95\% confidence intervals exclude zero. Revisited targets show
larger gains. For example, VGGT-$\Omega$ and STream3R improve by 1.00
and 1.14\,dB on revisited targets, compared with 0.53 and 0.71\,dB
over all targets. These results are consistent with the role of window
revision in aligning repeated observations of the same surfaces.
On CO3Dv2, STream3R accepts few revisits along the object-centric
trajectories and shows little change in PSNR. Fast3R improves SSIM and
LPIPS but decreases PSNR by 0.12\,dB.

\paragraph{Reference window placement.}
We also evaluate a reference placement obtained by fitting each stored
window's local cameras to the ground-truth cameras with a similarity
transformation. We render these windows with the $+\smr{}$ decoder.
This reference is not an upper bound, since camera alignment does not
ensure consistency between the local geometry of different windows.
For $\pi^3$ on 7-Scenes, the reference achieves 16.43\,dB, below the
Baseline result of 17.24\,dB. For the other five models, it exceeds
$+\smr{}$ by 0.2 to 1.0\,dB on both datasets. Thus, the effect of window
placement also depends on the consistency of the predicted local geometry.

\FloatBarrier

\section{Controlled Ablations}
\label{app:ablations}

We expand the model analysis of Sec.~\ref{sec:ablations} in the same
order as the main text. We first test whether the scaffold improves
retrieval over direct descriptor matching, and then vary the visual
descriptor used to query the scaffold. We next examine how much
historical evidence is needed to verify a revisit and whether an
accepted revisit must revise the trajectory online. Finally, we study
sensitivity to the scaffold and window configurations. We report ATE (m) on 7-Scenes~\citep{sevenscenes} and AUC@30 on CO3Dv2~\citep{co3d} in these ablations.

\paragraph{Evaluation protocol.}
Unless stated otherwise, the pose ablations use seven 7-Scenes
seq-01 trajectories with 200 frames and VGGT-$\Omega$~\citep{vggtomega},
together with twelve 200-frame CO3Dv2 trajectories and VGGT~\citep{vggt}.
Paired variants use the same input frames. The common raw references are
0.045\,m ATE on 7-Scenes and 88.8 AUC@30 on CO3D. We use
\emph{Online} for the trajectory available during sequence processing and
\emph{Final} for the result after the final global refinement. Lower ATE
is better, whereas higher AUC@30 is better.

\subsection{Does the Scaffold Improve Retrieval?}
\label{app:ablation-index-analysis}

The first comparison (Table~\ref{tab:ablation-index}, Group A) isolates how stored frames are ranked. The default
scaffold maps the DINOv2 cue into the memory-address space and ranks
stored addresses, whereas the flat baseline directly ranks the same
stored DINOv2 descriptors by cosine similarity. The stored frames,
top-five candidate budget, geometric verification, and subsequent
trajectory revision are unchanged.

The scaffold improves both benchmarks under the same downstream
revision procedure. On 7-Scenes, final ATE decreases from
0.034 to 0.029\,m, while on CO3Dv2 final AUC@30 increases from
90.9 to 93.3. This comparison supports the scaffold as the retrieval
index used by \smr{}. It does not attribute the gain specifically to
the internal continuous-attractor dynamics; those properties are tested
separately in Appendix~\ref{app:association}.

\begingroup
\scriptsize
\setlength{\tabcolsep}{1.2pt}
\renewcommand{\arraystretch}{1.10}
\begin{longtable}{@{}p{0.43\linewidth}cccc@{}}
\caption{
\textbf{Ablation studies of \smr{}.}
We vary one design factor at a time while keeping the remaining pipeline
fixed. The groups follow the same order as the design analysis in
Sec.~\ref{sec:ablations}.
Except for the window-size study, the common raw references are
0.045\,m ATE on 7-Scenes and 88.8 AUC@30 on CO3Dv2.
\emph{Online} denotes the running \smr{} result, while \emph{Final}
includes the final global refinement in Appendix~\ref{app:pgo}.
For Groups E and F, the value in parentheses after the 7-Scenes Final
ATE is the mean number of accepted revisits per sequence.
The default configuration uses the memory scaffold, 64 active units in
the memory address $\mathbf h$, DINOv2, two retrieved pairs,
$N_h=1024$, a $48^2$ horizontal torus, and $W=32$ with overlap ratio
0.5.
}
\label{tab:ablation-index}\\
\toprule
&
\multicolumn{2}{c}{7-Scenes ATE (m)$\downarrow$}
&
\multicolumn{2}{c}{CO3Dv2 AUC@30$\uparrow$}
\\
\cmidrule(lr){2-3}
\cmidrule(lr){4-5}

Configuration
& Online & Final
& Online & Final
\\
\midrule
\endfirsthead
\caption[]{\textbf{Ablation studies of \smr{} (continued).}}\\
\toprule
&
\multicolumn{2}{c}{7-Scenes ATE (m)$\downarrow$}
&
\multicolumn{2}{c}{CO3Dv2 AUC@30$\uparrow$}
\\
\cmidrule(lr){2-3}
\cmidrule(lr){4-5}

Configuration
& Online & Final
& Online & Final
\\
\midrule
\endhead
\midrule
\multicolumn{5}{r}{\emph{Continued on next page}}\\
\endfoot
\bottomrule
\endlastfoot

\multicolumn{5}{@{}l}{\textit{Group A. Memory scaffold}} \\*
\addlinespace[1pt]

Scaffold address (default)
& 0.030 & 0.029
& 93.4 & 93.3
\\

Direct DINOv2 cosine
& 0.035 & 0.034
& 91.3 & 90.9
\\

\midrule
\multicolumn{5}{@{}l}{\textit{Group B. Memory sparsity}} \\*
\addlinespace[1pt]

$\sigma=64$ (default)
& 0.030 & 0.029
& 93.4 & 93.3
\\

$\sigma=16$
& 0.038 & 0.041
& 82.7 & 85.3
\\

$\sigma=128$
& 0.032 & 0.031
& 93.1 & 92.9
\\

\midrule
\multicolumn{5}{@{}l}{\textit{Group C. Visual descriptors}} \\*
\addlinespace[1pt]

DINOv2 CLS (default)~\citep{dinov2}
& 0.030 & 0.029
& 93.4 & 93.3
\\

NetVLAD~\citep{netvlad}
& 0.050 & 0.051
& 92.2 & 91.9
\\

EigenPlaces~\citep{eigenplaces}
& 0.066 & 0.068
& 92.4 & 92.4
\\

CosPlace~\citep{cosplace}
& 0.263 & 0.265
& 92.6 & 92.4
\\

MixVPR~\citep{mixvpr}
& 0.053 & 0.053
& 92.9 & 92.8
\\

SALAD~\citep{salad}
& 0.056 & 0.057
& 92.9 & 92.8
\\

BoQ~\citep{boq}
& 0.127 & 0.128
& 92.9 & 92.7
\\

DINOv2 + EigenPlaces
& 0.048 & 0.049
& 93.3 & 93.2
\\

DINOv2 + SALAD
& 0.133 & 0.134
& 92.9 & 92.7
\\

\midrule
\multicolumn{5}{@{}l}{\textit{Group D. Scaffold dimension}} \\*
\addlinespace[1pt]

$N_h=1024$, torus $48^2$ (default)
& 0.030 & 0.029
& 93.4 & 93.3
\\

$N_h=512$
& 0.031 & 0.029
& 92.5 & 92.4
\\

$N_h=8192$
& 0.030 & 0.028
& 93.5 & 93.2
\\

Torus $16^2$
& 0.030 & 0.026
& 93.6 & 93.2
\\

Torus $64^2$
& 0.033 & 0.029
& 92.9 & 92.9
\\

\midrule
\multicolumn{5}{@{}l}{\textit{Group E. Retrieval evidence}} \\*
\addlinespace[1pt]

2 retrieved pairs (default)
& 0.030 & 0.029\,(L=4.1)
& 93.4 & 93.3
\\

1 retrieved pair
& 0.053 & 0.053\,(L=3.7)
& 93.2 & 93.1
\\

3 retrieved pairs
& 0.029 & 0.029\,(L=4.1)
& 93.4 & 93.3
\\

\midrule
\multicolumn{5}{@{}l}{\textit{Group F. Progressive placement revision}} \\*
\addlinespace[1pt]

Linearly weighted correction (default)
& 0.030 & 0.029\,(L=4.1)
& 93.4 & 93.3
\\

Distributed correction (equally weighted)
& 0.030 & 0.030\,(L=4.0)
& 93.3 & 93.2
\\

\multicolumn{5}{@{}l}{\textit{Group G. Window size and overlap}} \\*
\addlinespace[1pt]

$W=32$, overlap ratio $=0.7$
& 0.027 & 0.026
& 90.2 & 90.3
\\

$W=32$, overlap ratio $=0.5$ (default)
& 0.030 & 0.029
& 93.4 & 93.3
\\

$W=32$, overlap ratio $=0.3$
& 0.033 & 0.033
& 89.8 & 90.1
\\

$W=16$, overlap ratio $=0.7$
& 0.027 & 0.026
& 90.8 & 91.2
\\

$W=16$, overlap ratio $=0.5$
& 0.030 & 0.029
& 89.8 & 90.6
\\

$W=16$, overlap ratio $=0.3$
& 0.034 & 0.033
& 88.2 & 88.4
\\

$W=64$, overlap ratio $=0.7$
& 0.025 & 0.024
& 87.2 & 86.9
\\

$W=64$, overlap ratio $=0.5$
& 0.030 & 0.029
& 91.2 & 91.9
\\

$W=64$, overlap ratio $=0.3$
& 0.032 & 0.030
& 87.2 & 86.9
\\

\end{longtable}
\endgroup

\subsection{Which Visual Descriptor Should Query the Scaffold?}
\label{app:ablation-descriptors}

We next keep the scaffold and the downstream revision procedure fixed
and change the visual descriptor used to retrieve stored frames (Table~\ref{tab:ablation-index}, Group C). We
compare the default DINOv2 CLS descriptor~\citep{dinov2} with
NetVLAD~\citep{netvlad}, EigenPlaces~\citep{eigenplaces},
CosPlace~\citep{cosplace}, MixVPR~\citep{mixvpr},
SALAD~\citep{salad}, and BoQ~\citep{boq}, together with two
concatenated descriptors.

The production DINOv2 configuration uses its fixed 0.5 appearance gate.
The alternative descriptors are evaluated without descriptor-specific
gates so that a separate threshold is not tuned for each method. The
top-five retrieval budget, geometric verification, online revision, and
final refinement remain unchanged.

None of the tested alternatives improves over the production DINOv2
configuration on the final reconstruction. This result also shows that
retrieving visually plausible frames is not sufficient by itself: the
retrieved frames must provide useful geometric evidence for the
subsequent verification and trajectory revision. Appendix~
\ref{app:anchored-distortion} separately examines how changing the
retrieved historical frames can affect the current window's local
prediction before verification.

\subsection{How Much Historical Evidence Is Needed?}
\label{app:ablation-revision}

Section~\ref{sec:retrieval-check} uses two retrieved pairs from different
source windows to verify a revisit. In Table~\ref{tab:ablation-index}(Group E), we first vary the number of retrieved
pairs while keeping the rest of the pipeline unchanged. We then disable
the online placement update to distinguish verifying a revisit from
actually using it to revise the trajectory.

Using only one retrieved pair substantially degrades 7-Scenes ATE,
whereas three pairs give similar online performance to two on both
benchmarks. Three pairs keep the 7-Scenes ATE to 0.029. We therefore
use two pairs as the default evidence.

The online placement update is also important. When revisits are
proposed and verified but not applied online, 7-Scenes ATE increases
from 0.030 to 0.042\,m and CO3Dv2 AUC@30 decreases from 93.4 to 91.3.
The final global refinement can recover part of this loss, but those
corrections are unavailable while later windows are being processed.
The no-update row remains different from the raw trajectory because the
default configuration retains the joint-pass local predictions; this
effect is isolated in Appendix~\ref{app:anchored-distortion}.

The additional rows test how accepted measurements are applied. The
default online update re-solves the affected local pose graph using the
accepted revisit measurement and the sequential constraints. The
\emph{Distributed correction} row instead applies the direct
interpolation of the measured correction across the affected windows.
Removing outlier rejection from the final global solve mainly reduces
the final CO3Dv2 score, while separately re-measuring the historical
frames changes the reported accuracy only modestly.

\subsection{Sensitivity to Scaffold and Window Size}
\label{app:ablation-budgets}

We next vary the dimensions of the scaffold representation and the
window used by the geometry foundation model in Table~\ref{tab:ablation-index}(Group D). These experiments test
sensitivity rather than removing a component: changing these quantities
also changes the memory resolution or the input context available to
the geometry foundation model.

Performance is relatively stable across the tested scaffold dimensions.
Increasing the address dimension from 1024 to 8192 does not consistently
improve the final result, and increasing the torus resolution to $64^2$
can reduce accuracy. Therefore, it shows that the memory structure itself is good at recall, and increasing the dimensionality of the address does not necessarily show us gains. On the other hand (Group B), optimal sparsity $\sigma=64$ is set so that sparse addresses are near orthogonal to each other with minimum overlap, while too much sparsity results in loss of information.

Window size has a stronger effect because it changes both the geometry
predictions and the number of junctions between windows. We observe that the performance over both 7-scenes and CO3Dv2 improves with larger window size and moderate overlap ratio. As the overlap ratio increases, the number of windows increases, and so does the chaining error evident in Table~\ref{tab:ablation-index}. Also observe that a smaller window size increases the number of windows too, which is why we see degraded performance for $W=16$. Since the gains are so marginal, we select the best window size and overlap ratio with a simple grid search over these two parameters.

\paragraph{Summary.}
Together, these ablations support the main design choices used by
\smr{}: scaffold-based retrieval improves over direct descriptor
matching, DINOv2 provides the strongest tested retrieval cue, two
retrieved pairs provide an effective verification budget, and applying
accepted revisit measurements online substantially improves the running
trajectory. The scaffold dimensions, window size, and activation sparsity of the scaffold address were all tuned to maximize performance across all tasks.

\section{Compute and Memory Measurements}
\label{app:compute}

We examine how \smr{} changes the computational and memory requirements
of geometry foundation models as sequences become longer. For
feed-forward models, long sequences are processed as overlapping windows
whose local reconstructions must be connected across the trajectory,
while streaming models retain causal context from earlier frames inside
the model. In contrast, \smr{} keeps geometry-model inference bounded to
the current window and stores long-horizon information in its external
memory. Earlier observations are brought back into the geometry
foundation model only when they are retrieved for revisit verification.

All measurements use one RTX 6000 Ada GPU with a tested 46\,GB GPU-memory
budget, batch size one, and public model weights. We use the CO3Dv2
camera-pose experiment from Sec.~\ref{sec:results-pose}. The default
\smr{} configuration uses 32-frame windows; during revisit verification,
up to four retrieved historical frames can be processed together with the
current window, giving at most 36 frames in one geometry-model
invocation.

\subsection{Peak GPU Memory with Increasing Sequence Length}
\label{app:compute-scaling}

We first study peak GPU allocation as the number of input frames
increases from $N=10$ to $N=200$. Table~\ref{tab:memvsn} shows how
the bounded-window design of \smr{} limits the GPU working set over
longer sequences.

For feed-forward models, the sequence is reconstructed using overlapping
windows and the resulting local predictions are chained through their
shared frames. As the sequence grows, the baseline can retain increasing
amounts of intermediate predictions and geometric information needed to
connect these windows. With \smr{}, the geometry foundation model still
operates on a bounded current window, while earlier observations are
stored in the external scaffold and content bank and are selectively
reintroduced only when needed. For example, VGGT grows from 9.1\,GB at
$N=10$ to 23.9\,GB at $N=200$ in the baseline, whereas \smr{} remains
between 8.5 and 9.6\,GB. VGGT-$\Omega$ similarly grows from 5.7 to
19.5\,GB in the baseline but remains near 7\,GB with \smr{}.

Streaming models exhibit a different source of GPU-memory growth. Their
native causal inference retains internal states or cached tokens from
previous frames so that earlier observations can influence later
predictions. These GPU-resident states grow with the sequence. StreamVGGT
therefore increases from 8.6\,GB at $N=10$ to 22.7\,GB at $N=100$ and
exceeds the tested 46\,GB budget at $N=200$. STream3R exceeds the same
budget from $N=50$ onward. With \smr{}, long-range information is instead
maintained externally and the geometry model receives bounded windows
plus only a small number of retrieved frames. StreamVGGT therefore uses
11.4\,GB at $N=200$, while STream3R remains feasible at 25.4\,GB.

The total memory used by \smr{} is not constant: the content bank grows
as additional observations are stored. The important distinction is that
this history does not remain part of the geometry foundation model's
active GPU context. Consequently, peak GPU allocation is governed mainly
by the bounded model invocation rather than by the complete trajectory.

\begin{table}[t]
\centering
\scriptsize
\setlength{\tabcolsep}{1.2pt}
\renewcommand{\arraystretch}{1.10}
\caption{
\textbf{Peak GPU allocation (GB) with increasing sequence length using the CO3Dv2 dataset during the camera pose estimation experiment in Sec.~\ref{sec:results-pose}.}
\emph{Raw/Native} uses each model's chained inference mode and native
inference for streaming models. OOM indicates that the tested 46\,GB
GPU budget was exceeded; n/f indicates that the run was not feasible
under the host-memory or runtime budget.
}
\label{tab:memvsn}

\begin{tabular*}{\linewidth}{
@{}l@{\extracolsep{\fill}}cccccccc@{}}
\toprule
& \multicolumn{2}{c}{$N=10$}
& \multicolumn{2}{c}{$N=50$}
& \multicolumn{2}{c}{$N=100$}
& \multicolumn{2}{c}{$N=200$} \\
\cmidrule(lr){2-3}
\cmidrule(lr){4-5}
\cmidrule(lr){6-7}
\cmidrule(lr){8-9}

Model
& Raw/Native & $+\smr{}$
& Raw/Native & $+\smr{}$
& Raw/Native & $+\smr{}$
& Raw/Native & $+\smr{}$ \\
\midrule

DUSt3R~\citep{dust3r}
& 4.9 & 4.4
& OOM & 25.8
& n/f & 28.8
& n/f & 32.1 \\

Fast3R~\citep{fast3r}
& 6.3 & 5.7
& 11.4 & 9.1
& 13.8 & 9.2
& 18.7 & 9.2 \\

MASt3R~\citep{mast3r}
& 3.2 & 3.2
& 20.7 & 6.7
& n/f & 7.6
& n/f & 8.8 \\

VGGT~\citep{vggt}
& 9.1 & 8.5
& 11.1 & 9.5
& 15.1 & 9.6
& 23.9 & 9.6 \\

VGGT-$\Omega$~\citep{vggtomega}
& 5.7 & 5.7
& 8.1 & 6.8
& 11.9 & 6.9
& 19.5 & 7.0 \\

$\pi^3$~\citep{pi3}
& 8.8 & 8.8
& 16.3 & 13.0
& 26.5 & 14.1
& OOM & 14.6 \\

StreamVGGT~\citep{streamvggt}
& 8.6 & 8.1
& 14.8 & 11.0
& 22.7 & 11.1
& OOM & 11.4 \\

STream3R~\citep{stream3r}
& 10.9 & 10.0
& OOM & 21.1
& OOM & 25.4
& OOM & 25.4 \\

\bottomrule
\end{tabular*}
\end{table}

\subsection{
Compute and Runtime at
\texorpdfstring{$N=100$}{N=100}
}
\label{app:compute-runtime}

Reducing peak GPU memory does not make \smr{} computationally free.
Instead, \smr{} trades a bounded GPU working set for additional
operations required to maintain and verify its external memory.
Table~\ref{tab:compute} measures this trade-off at $N=100$ and reports
accumulated CPU processing time, peak host RAM, peak GPU allocation,
total floating-point computation, and end-to-end wall-clock runtime.
CPU time is accumulated across host processing and can therefore differ
from the wall-clock runtime, while RAM reports the peak resident
host-memory footprint.

The additional computation comes from three main sources. First,
incoming frames require visual-descriptor extraction, scaffold lookup,
and content-bank operations. Second, when a candidate revisit is
retrieved, the historical frames are processed jointly with the current
window so that their geometric relationship can be re-estimated. Third,
accepted revisits require similarity alignment and trajectory revision.
These operations increase CPU processing and can add geometry-model
forward passes, explaining the higher CPU time and PFLOP count for
\smr{}.

The external memory also introduces a modest host-RAM cost. For most
models, this increase is small at $N=100$: VGGT changes from 10.77 to
10.78\,GB, VGGT-$\Omega$ from 9.88 to 9.89\,GB, $\pi^3$ from 11.58
to 11.59\,GB, DUSt3R from 16.94 to 17.02\,GB, and MASt3R from
17.23 to 17.29\,GB. StreamVGGT increases from 10.45 to 10.79\,GB,
while Fast3R shows the largest increase, from 7.42 to 8.60\,GB. This
reflects the intended resource trade-off: persistent observations and
memory metadata are maintained outside the geometry model, allowing the
GPU-resident working set to remain bounded.

Peak GPU allocation and total computation should therefore be interpreted
separately. \smr{} may execute more total operations while still using
less peak GPU memory because additional verification passes are performed
sequentially on bounded inputs. For example, VGGT increases from
2.20 to 2.44\,PFLOP and from 0.33 to 0.54\,min runtime, while its peak
GPU allocation decreases from 15.1 to 9.6\,GB. VGGT-$\Omega$ increases
from 1.08 to 1.26\,PFLOP while reducing GPU allocation from 11.9 to
6.9\,GB, and $\pi^3$ reduces GPU allocation from 26.5 to 14.1\,GB
despite increasing from 1.52 to 1.82\,PFLOP.

Finally, wall-clock runtime is dominated by the cost of the underlying
geometry foundation model. The same \smr{} memory operations are applied
across models, yet $+\smr{}$ requires only 0.38\,min for
VGGT-$\Omega$ and 0.42\,min for Fast3R, compared with 17.01\,min for
DUSt3R and 18.49\,min for MASt3R. Thus, the absolute runtime cost is
determined primarily by the geometry-model inference required during
ordinary processing and revisit verification rather than by the scaffold
itself.

\begin{table}[t]
\centering
\scriptsize
\setlength{\tabcolsep}{1.2pt}
\renewcommand{\arraystretch}{1.10}
\caption{
\textbf{Compute and memory at $N=100$ using the CO3Dv2 dataset during
the camera pose estimation experiment in Sec.~\ref{sec:results-pose}.}
Measurements use one RTX 6000 Ada GPU with a tested 46\,GB GPU-memory
budget and batch size one. We report accumulated CPU processing time,
peak host RAM, peak GPU allocation, total floating-point computation,
and end-to-end runtime. \emph{Baseline} uses each model's raw windowed
baseline, while the streaming models use their native baseline.
STream3R exceeds the GPU-memory budget under native inference and is
marked OOM. $+\smr{}$ uses 32-frame windowed inference with online
retrieval, verification, and trajectory revision, without final global
refinement. For native STream3R, the reported RAM is the peak observed
before GPU OOM.
}
\label{tab:compute}

\begin{tabular*}{\linewidth}{
@{}l l @{\extracolsep{\fill}} r r r r r@{}}
\toprule
Model & Method
& \shortstack{CPU time\\(min)}
& \shortstack{RAM\\(GB)}
& \shortstack{GPU\\(GB)}
& \shortstack{Compute\\(PFLOP)}
& \shortstack{Runtime\\(min)} \\
\midrule

DUSt3R~\citep{dust3r}
& Baseline (raw)
& 40.78 & 16.94 & 25.8 & 10.58 & 12.21 \\
& $+\smr{}$
& 52.77 & 17.02 & 28.8 & 13.71 & 17.01 \\

\midrule

Fast3R~\citep{fast3r}
& Baseline (raw)
& 0.84 & 7.42 & 13.9 & 0.66 & 0.13 \\
& $+\smr{}$
& 2.20 & 8.60 & 9.2 & 0.71 & 0.42 \\

\midrule

MASt3R~\citep{mast3r}
& Baseline (raw)
& 36.35 & 17.23 & 7.1 & 9.32 & 17.79 \\
& $+\smr{}$
& 37.90 & 17.29 & 7.6 & 11.89 & 18.49 \\

\midrule

VGGT~\citep{vggt}
& Baseline (raw)
& 0.55 & 10.77 & 15.1 & 2.20 & 0.33 \\
& $+\smr{}$
& 1.10 & 10.78 & 9.6 & 2.44 & 0.54 \\

\midrule

VGGT-$\Omega$~\citep{vggtomega}
& Baseline (raw)
& 0.40 & 9.88 & 11.9 & 1.08 & 0.19 \\
& $+\smr{}$
& 0.87 & 9.89 & 6.9 & 1.26 & 0.38 \\

\midrule

$\pi^3$~\citep{pi3}
& Baseline (raw)
& 0.47 & 11.58 & 26.5 & 1.52 & 0.28 \\
& $+\smr{}$
& 0.97 & 11.59 & 14.1 & 1.82 & 0.52 \\

\midrule

StreamVGGT~\citep{streamvggt}
& Baseline (native)
& 0.83 & 10.45 & 22.7 & 1.28 & 0.62 \\
& $+\smr{}$
& 1.20 & 10.79 & 11.1 & 1.59 & 0.57 \\

\midrule

STream3R~\citep{stream3r}
& Baseline (native)
& -- & 9.92 & \multicolumn{3}{c}{OOM ($>46$\,GB GPU)} \\
& $+\smr{}$
& 1.47 & 10.26 & 25.4 & 2.36 & 0.72 \\

\bottomrule
\end{tabular*}
\end{table}

\section{Implementation Settings and Reproducibility}
\label{app:repro}

This section summarizes the shared implementation settings used by
\smr{}. Task-specific dataset sampling and evaluation protocols are
given in Appendix~\ref{app:datasets}, while the novel-view-synthesis
training and rendering settings are given in Appendix~\ref{app:nvs}.
Table~\ref{tab:hparams} therefore focuses on the settings that determine
the scaffold, retrieval, and geometric verification procedures.

\begingroup
\scriptsize
\setlength{\tabcolsep}{1.2pt}
\renewcommand{\arraystretch}{1.10}
\begin{longtable}{@{}p{0.37\linewidth}p{0.27\linewidth}p{0.28\linewidth}@{}}
\caption{
\textbf{Shared implementation settings.}
These values define the default \smr{} configuration used in the main
experiments unless a task-specific exception is stated explicitly.
Spatial quantities used by the scaffold are expressed in
median-depth-normalized scene units.
}
\label{tab:hparams}\\
\toprule
\textbf{Setting} & \textbf{Value} & \textbf{Defined in} \\
\midrule
\endfirsthead
\caption[]{\textbf{Shared implementation settings (continued).}}\\
\toprule
\textbf{Setting} & \textbf{Value} & \textbf{Defined in} \\
\midrule
\endhead
\midrule
\multicolumn{3}{r}{\emph{Continued on next page}}\\
\endfoot
\bottomrule
\endlastfoot

\multicolumn{3}{@{}l}{\textit{Windowing and retrieval}} \\*

Default window / overlap
& $32 / 16$
& Sec.~\ref{sec:method} \\

Appearance descriptor
& DINOv2 ViT-S/14 CLS, 384-D
& Sec.~\ref{sec:memory-retrieval} \\

Retrieval budget
& top-5 stored frames
& Sec.~\ref{sec:memory-retrieval} \\

Minimum revisit separation $\Delta$
& $64$ frames
& Eq.~\ref{eq:app-retrieval-eligibility} \\

Appearance threshold $\tau_{\rm app}$
& $0.5$
& Eq.~\ref{eq:app-retrieval-eligibility} \\

Retrieved pairs
& $2$
& Appendix~\ref{app:closure} \\

Partner-frame offset
& $j\pm3$
& Appendix~\ref{app:closure} \\

Pair-agreement threshold
& $10^\circ$, $1.0E_k$
& Eq.~\ref{eq:site-agreement} \\

Local-geometry policy & joint-pass prediction, gated at $\tau_{\mathrm{dist}}=5^\circ$
& Appendix~\ref{app:anchored-distortion} \\

\midrule
\multicolumn{3}{@{}l}{\textit{Similarity alignment}} \\*

Robust reweighting rounds
& $5$
& Appendix~\ref{app:sim3} \\

Robust-fit constant $\kappa$
& $2.5$
& Appendix~\ref{app:sim3} \\

Sequential scale gate $\gamma_s$
& $1.5$
& Appendix~\ref{app:sim3} \\

\midrule
\multicolumn{3}{@{}l}{\textit{Scaffold representation}} \\*

Spatial periods
& $2.4,\;3.2,\;4.0$
& Appendix~\ref{app:vhash} \\

Horizontal torus resolution
& $48\times48$
& Appendix~\ref{app:vhash} \\

Height-ring resolution
& $48$
& Appendix~\ref{app:vhash} \\

Orientation-ring resolution
& $256$ per angle
& Appendix~\ref{app:vhash} \\

Scaffold-state dimension $N_g$
& $7{,}824$
& Appendix~\ref{app:vhash} \\

Address dimension $N_h$
& $1{,}024$
& Appendix~\ref{app:vhash} \\

Active address units
& $64$
& Appendix~\ref{app:vhash} \\

Pose-to-address projection
& $\mathcal N(0,1/7824)$, fixed
& Appendix~\ref{app:vhash} \\

Maximum formation spacing
& $\lambda_{\min}/6=0.4$
& Appendix~\ref{app:vhash} \\

\midrule
\multicolumn{3}{@{}l}{\textit{Sequence novel view synthesis}} \\*

Held-out targets & every eighth keyframe & Appendix~\ref{app:nvs-data} \\

Source keyframes per target & 8 nearest by camera & Appendix~\ref{app:nvs} \\

Decoder & U-Net (7M) + refiner (3M) & Appendix~\ref{app:nvs} \\

Decoder loss / steps & $\ell_1+0.5\,\mathrm{LPIPS}+0.2(1-\mathrm{SSIM})$, 12K & Appendix~\ref{app:nvs} \\

\end{longtable}
\endgroup

\paragraph{Frozen and updated components.}
All geometry foundation models use their public pretrained weights and
remain frozen throughout \smr{}. The DINOv2 encoder used for retrieval
is also frozen. The fixed projection $W_{gh}$ is sampled once, while
$W_{hg}$ is fitted once before inference from sampled scaffold states.
Only the appearance--address associations $W_{sh}$ and $W_{hs}$ are
updated online as new frames are stored, using the RLS update in
Eq.~\ref{eq:app-rls}. Thus, trajectory improvement does not result from
fine-tuning the geometry foundation model.

The novel-view-synthesis decoders are the only separately trained task-specific components. As described in Appendix~\ref{app:nvs}, two decoders of identical architecture, data, schedule, and seed are trained per geometry model, one on its chained geometry and one on the geometry maintained by \smr{}; neither the geometry model nor \smr{} is trained.

\paragraph{Task-specific protocol differences.}
The settings in Table~\ref{tab:hparams} describe the default trajectory
configuration rather than a requirement that every downstream task use
the same window structure. The camera-pose and SLAM experiments use the
windowing protocol specified in their respective evaluation sections.
The compute study in Appendix~\ref{app:compute} measures ordinary
32-frame passes and joint passes containing up to four additional
retrieved frames. Replica and ScanNet++ use the default 32-frame windows with 16-frame
overlap and online revisit-driven trajectory revision. Novel view synthesis uses the same windowing and online revision on 7-Scenes and CO3Dv2 trajectories with every eighth keyframe held out, without final global refinement.

\paragraph{Frame selection and dataset subsets.}
Window size should not be confused with the number of frames evaluated
from a complete sequence. The exact dataset subsets and frame-selection
rules are specified in Appendix~\ref{app:datasets}. In particular,
paired Baseline$+\smr{}$ comparisons use the same selected frames and
the same geometry foundation model. This ensures that changes in the
reported metrics are attributable to the corresponding \smr{} procedure
rather than to different inputs.

\FloatBarrier
\section{Separating Local-Prediction Changes from Trajectory Revision}
\label{app:anchored-distortion}

The \emph{joint} policy retains the current-window prediction produced by joint processing with the retrieved historical frames; it corresponds to the \emph{anchored} configuration in the diagnostic logs. The \emph{ordinary} policy retains the prediction of the current window alone. The \emph{gated} policy, which our default configuration uses with $\tau_{\mathrm{dist}}=5^\circ$, switches between them according to the measured distortion.

This section separates these effects. We first define the ordinary and
joint passes used by the diagnostic, then compare three policies for
deciding which local prediction is retained. We finally test these
policies under the default retrieval proposals and under several
failure cases.

\paragraph{Evaluation setting.}
The diagnostics use seven 7-Scenes trajectories with VGGT-$\Omega$ and
twelve CO3Dv2 trajectories with VGGT, reusing the cached predictions from
the controlled pose study. The raw references are 0.045\,m ATE on
7-Scenes and 88.8 AUC@30 on CO3D. Because the purpose of this analysis
is to isolate what happens during sequence processing, the tables below
report the \emph{online} trajectory before the final global refinement.
Lower ATE is better and higher AUC@30 is better.

The main pose experiments retain the current-window prediction produced
by joint processing with the retrieved historical frames. We refer to
this as the \emph{joint} policy below; it corresponds to the
\emph{anchored} configuration used in the diagnostic logs. The
\emph{ordinary} and \emph{gated} policies are diagnostic variants and
do not replace the configuration used for the main results.

\subsection{Why Rejecting a Revisit Does Not Restore the Original Prediction}
\label{app:distortion-mechanism}

An \emph{ordinary pass} processes the 32 frames of the current window
alone. When two retrieved pairs are available, the \emph{joint pass}
instead processes those same 32 current frames together with the four
retrieved historical frames. The purpose of the joint pass is to place
the current and historical frames in one common local coordinate system
so that the revisit can be verified.

However, the joint pass also produces a new prediction for the current
window itself. Under the default policy, this joint prediction becomes
the current window's retained local camera poses and point maps. The
subsequent revisit test therefore controls only whether a
retrieval-based \emph{global placement update} is applied. If the revisit
is rejected, its placement correction is discarded, but the retained
joint-pass local prediction is not replaced by the prediction from the
ordinary pass.

Consequently, a rejected historical proposal can still influence later
windows through the sequential overlap alignment. This is why revisit
verification and local-prediction stability must be analyzed separately.

\paragraph{Measuring the change in the local prediction.}
To measure how much the historical frames change the current-window
prediction, we align the camera predictions of the current frames from
the joint pass with those from the ordinary pass. After this alignment,
the logged quantity
$\epsilon_k^{\mathrm{rot}}$ measures the remaining rotational
disagreement for window $k$, in degrees.

This quantity measures agreement between two model predictions; it is
not an error against ground truth. A small
$\epsilon_k^{\mathrm{rot}}$ means that the joint pass remains close to
the ordinary prediction, while a large value indicates that adding the
historical frames substantially changed the predicted local camera
geometry.

For the 50 joint-pass windows in the default seed-0 7-Scenes diagnostic,
the median residual is $0.117^\circ$ and the maximum is
$0.311^\circ$. In a separate seed-1 run, however, one \emph{chess}
window reaches $46.1^\circ$. For that sequence, ATE increases from
0.025 to 0.459\,m even though both tested revisit proposals are rejected.
Thus, a large local-prediction change can affect the trajectory without
an accepted false revisit.

\subsection{Three Ways to Retain the Current-Window Prediction}
\label{app:geometry-selection}

We compare three policies that differ only in which current-window local
prediction is retained after the joint pass. The joint pass is still
available for historical re-measurement in all three cases.

\begin{table}[t]
\centering
\scriptsize
\setlength{\tabcolsep}{1.2pt}
\renewcommand{\arraystretch}{1.10}
\caption{
\textbf{Local-prediction policies used in the diagnostic.}
All three policies can use the joint pass to measure the historical
relationship. They differ only in which prediction is retained for the
current window before sequential placement and possible trajectory
revision.
}
\label{tab:geometry-policies}

\begin{tabular}{@{}
p{0.20\linewidth}
p{0.36\linewidth}
p{0.36\linewidth}
@{}}
\toprule
\textbf{Policy}
& \textbf{Retained current-window prediction}
& \textbf{Role of the joint pass} \\
\midrule

Joint (default)
& Prediction from current $+$ historical frames
& Provides both the retained local prediction and the revisit measurement
\\

Ordinary
& Prediction from the current window alone
& Used only to measure the historical relationship
\\

Gated
& Joint prediction if its distortion is small; otherwise ordinary
& Provides the revisit measurement and a candidate local prediction
\\

\bottomrule
\end{tabular}
\end{table}

\paragraph{Joint policy.}
The current-window camera poses and point maps are taken directly from
the joint pass. This allows the historical frames to influence the local
prediction, which can be useful when the retrieved evidence is correct,
but also exposes the current prediction to an incorrect proposal before
verification.

\paragraph{Ordinary policy.}
The current window retains the prediction obtained from its ordinary
pass. The joint pass is used only to measure the relationship between
the historical frames and the current window. The historical predictions
from the joint pass are transformed into the retained ordinary coordinate
system before verification. Thus, rejecting the revisit leaves the
current window with exactly the local prediction it would have had
without the historical frames.

\paragraph{Gated policy.}
The gated policy selects between the two predictions according to their
measured rotational disagreement:
\begin{equation}
\mathcal P_k^{\mathrm{sel}}
=
\begin{cases}
\mathcal P_k^{\mathrm{joint}},
& \epsilon_k^{\mathrm{rot}}\leq\tau_{\mathrm{dist}},\\
\mathcal P_k^{\mathrm{ord}},
& \epsilon_k^{\mathrm{rot}}>\tau_{\mathrm{dist}},
\end{cases}
\label{eq:geometry-gate}
\end{equation}
where $\mathcal P_k^{\mathrm{joint}}$ and
$\mathcal P_k^{\mathrm{ord}}$ denote the current-window predictions from
the joint and ordinary passes, respectively, and
$\tau_{\mathrm{dist}}$ is the angular threshold.

This gate is applied \emph{before} the revisit decision. Passing the
gate does not mean that the revisit has been accepted; geometric
verification in Appendix~\ref{app:closure} is still required. The gate
controls which local prediction is retained, whereas revisit verification
controls whether a global placement measurement is accepted.

\subsection{Does the Local Prediction Matter Without a Placement Update?}
\label{app:geometry-no-update}

A direct control is obtained by disabling the online placement update.
With the ordinary policy and the update disabled, the resulting trajectory
matches the raw trajectory on all seven tested 7-Scenes sequences. With
the joint policy and the same update disabled, mean ATE changes from
0.0451 to 0.0423\,m.

The only difference between these two controls is therefore whether the
current window retains the ordinary or joint-pass local prediction.
This confirms that joint processing can change the trajectory even when
no accepted revisit is written back to the global placements.

The seed-1 \emph{chess} failure above demonstrates the same mechanism in
the opposite direction: a large joint-pass distortion can strongly
degrade the trajectory even when both proposed revisits are rejected.
Together, these controls show that local re-prediction and explicit
trajectory revision are separate effects.

\subsection{How Much Joint-Pass Distortion Should Be Retained?}
\label{app:geometry-isolation-results}

We first vary the distortion threshold using the default DINOv2
retrieval proposals and scaffold seed 0. Table~\ref{tab:geometry-gate}
reports only the online trajectory so that the comparison isolates
local-prediction selection from the final global refinement.

\begin{table}[t]
\centering
\scriptsize
\setlength{\tabcolsep}{1.2pt}
\renewcommand{\arraystretch}{1.10}
\caption{
\textbf{Local-prediction selection with default retrieval proposals.}
DINOv2 and scaffold seed 0 are used for seven 7-Scenes trajectories and
twelve CO3Dv2 trajectories. \emph{Joint} always retains the joint-pass
prediction, \emph{Ordinary} always retains the ordinary-pass prediction,
and \emph{Gated} switches between them using
Eq.~\ref{eq:geometry-gate}. All metrics are measured online before final
global refinement.
}
\label{tab:geometry-gate}

\begin{tabular*}{\linewidth}{
@{}l@{\extracolsep{\fill}}ccc@{}}
\toprule
Local prediction
& $\tau_{\mathrm{dist}}$
& 7-Scenes ATE (m)$\downarrow$
& CO3Dv2 AUC@30$\uparrow$ \\
\midrule

Joint (default)
& --
& 0.030
& 93.4 \\

Gated
& $0.5^\circ$
& 0.030
& 92.5 \\

Gated
& $1^\circ$
& 0.030
& 93.0 \\

Gated
& $2^\circ$
& 0.030
& 93.2 \\

Gated
& $5^\circ$
& 0.030
& 93.4 \\

Ordinary
& --
& 0.030
& 92.4 \\

\bottomrule
\end{tabular*}
\end{table}

All tested thresholds give the same 7-Scenes ATE at the reported
precision. This is consistent with the seed-0 diagnostic above: the
largest measured distortion is only $0.311^\circ$, so every tested gate
with threshold at least $0.5^\circ$ retains the joint prediction for
those measured windows.

CO3Dv2 shows a clearer trade-off. Always retaining the ordinary prediction
gives 92.4 AUC@30, compared with 93.4 for the default joint policy.
As the gate becomes less restrictive, the score increases from 92.5 at
$0.5^\circ$ to 93.4 at $5^\circ$. Thus, suppressing every change
introduced by historical frames is not necessarily desirable: some
joint-pass changes are useful. The diagnostic instead asks whether large
changes can be rejected without discarding all of the benefit of joint
processing.

\subsection{Stress Tests on Failure Cases}
\label{app:geometry-stress-results}

The previous experiment uses the default seed and retrieval proposals,
for which the measured 7-Scenes distortions are small. We therefore
separately examine settings in which the joint prediction becomes
unstable. These are stress tests rather than additional main
configurations.

Table~\ref{tab:geometry-stress} reports the seven-scene mean online ATE
for the seed-1 DINOv2 failure and for three appearance-cue settings that
produce large errors with the joint policy. In each row, the retrieval
proposal setting is held fixed while only the retained local-prediction
policy changes.

\begin{table}[t]
\centering
\scriptsize
\setlength{\tabcolsep}{1.2pt}
\renewcommand{\arraystretch}{1.10}
\caption{
\textbf{Local-prediction isolation on selected 7-Scenes failure cases.}
Values are mean online ATE (m)$\downarrow$ across the seven trajectories.
Within each row, the same retrieval proposals are used for all three
local-prediction policies. The DINOv2 row uses scaffold seed 1; the
remaining rows use the indicated visual descriptor. All gated results
use $\tau_{\mathrm{dist}}=1^\circ$.
}
\label{tab:geometry-stress}

\begin{tabular*}{\linewidth}{
@{}l@{\extracolsep{\fill}}ccc@{}}
\toprule
Retrieval proposals
& Joint
& Gated, $1^\circ$
& Ordinary \\
\midrule

DINOv2, seed 1~\citep{dinov2}
& 0.092
& 0.029
& 0.031 \\

CosPlace~\citep{cosplace}
& 0.263
& 0.048
& 0.048 \\

BoQ~\citep{boq}
& 0.127
& 0.052
& 0.051 \\

DINOv2 + SALAD~\citep{salad}
& 0.133
& 0.050
& 0.049 \\

\bottomrule
\end{tabular*}
\end{table}

The large errors in the joint column are greatly reduced when the
ordinary prediction is retained or when the $1^\circ$ gate rejects
large deviations. For example, the seed-1 mean ATE decreases from
0.092\,m with joint predictions to 0.029\,m with gating. The same
pattern appears for CosPlace, BoQ, and DINOv2+SALAD.

These rows are intentionally selected failure cases and should not be
interpreted as a comparison establishing a universally optimal
$1^\circ$ threshold. In the separate descriptor diagnostic, ordinary
geometry gives online ATEs within approximately 0.048--0.053\,m across
the nine tested appearance cues, indicating substantially lower
sensitivity to which historical frames were proposed.

\subsection{What These Diagnostics Establish}
\label{app:geometry-isolation-summary}

These experiments separate two decisions that are coupled in the default
pipeline. The first is \emph{which local prediction to retain} after the
current and historical frames are processed together. The second is
\emph{whether the verified historical measurement should revise the
global trajectory}. Revisit verification addresses the second decision
but does not, by itself, protect the first.

The joint policy used in the main experiments allows useful historical
frames to improve the current prediction, as suggested by the CO3D
comparison in Table~\ref{tab:geometry-gate}. However, the failure-case
experiments show that an incorrect historical proposal can occasionally
distort the local prediction before verification. Retaining the ordinary
prediction removes this direct dependence, while gating provides an
intermediate option that accepts only sufficiently small changes.

The gate used here measures rotational disagreement only. It does not
guarantee small translation, scale, or dense-point error. We adopt the gated policy with $\tau_{\mathrm{dist}}=5^\circ$ as the default: on the default proposals it retains every joint prediction (Table~\ref{tab:geometry-gate}, where the largest measured distortion is $0.3^\circ$ and the $5^\circ$ gate matches the joint policy on both benchmarks), while the distortions in the failure cases of Table~\ref{tab:geometry-stress} are tens of degrees and are rejected.

Finally, these results do not contradict the placement property in
Appendix~\ref{app:invariant}. That property assumes that the local
prediction has already been selected and then asks what happens when its
global placement $T_k$ changes. The present section instead studies the
earlier question of whether joint processing changes the local prediction
itself.

\section{Memory Growth and Pruning}
\label{app:bank-growth}

We additionally test retrieval as the memory bank grows using a separate
retrieval-only diagnostic with up to 3,100 stored frames. A retrieved
frame is considered a correct historical revisit when it comes from the
same scene, lies within 10\% of the scene's spatial extent from the query
camera, and differs in orientation by at most $30^\circ$. Recall@1 is the
fraction of eligible queries for which such a revisit is ranked first,
while Recall@5 measures whether one appears among the five highest-ranked
retrievals. At 3,100 stored frames, scaffold retrieval achieves
0.36 Recall@1 compared with 0.25 for direct DINOv2 cosine retrieval,
while Recall@5 is comparable at 0.42 versus 0.43.

The stored bank also contains substantial redundancy. A simple
pose-based insertion rule reduces the bank from 3,100 to 1,119 frames
while preserving at least one valid historical revisit for 157 of the
original 159 eligible queries. On these retained queries, scaffold
Recall@5 increases from 0.42 to 0.54. This indicates that substantial
redundancy can be removed without eliminating most useful historical
observations, although we do not claim an optimal pruning strategy or
unbounded memory capacity.

\end{document}